\documentclass[10pt,twocolumn,letterpaper]{article}
\pdfoutput=1
\usepackage[pagebackref=true,breaklinks=true,colorlinks,bookmarks=false]{hyperref}

\usepackage{times}
\usepackage{epsfig}
\usepackage{graphicx}
\usepackage{amsmath}
\usepackage{amsfonts}
\usepackage{amssymb}
\usepackage{nicefrac}
\usepackage{booktabs}
\usepackage{mathtools}

\usepackage{verbatim}
\usepackage{flushend}
\usepackage[T1]{fontenc}           
\usepackage[utf8]{inputenc}        
\usepackage{bm}                    
\usepackage{paralist}              
\usepackage[capitalise]{cleveref}  

\usepackage{pgf}
\usepackage{tikz}
\usepackage{pgfplots}
\usetikzlibrary{plotmarks}
\usetikzlibrary{shapes}
\usepgfplotslibrary{units}
\pgfplotsset{compat=newest}
\pgfplotsset{plot coordinates/math parser=false}
\pgfplotsset{try min ticks=3}
\pgfplotsset{max space between ticks=20pt}
\pgfplotsset{invoke before crossref tikzpicture={\tikzexternaldisable},invoke
  after crossref tikzpicture={\tikzexternalenable}}
\pgfplotsset{every x tick label/.append style={font=\footnotesize, yshift=0.5ex}}
\pgfplotsset{every y tick label/.append style={font=\footnotesize, xshift=0.0ex}}

\newlength\fwidth%
\newlength\fheight%

\usepackage[font+=small]{subcaption}

\newcommand{\T}{^{\mathstrut\scriptscriptstyle{\top}}}
\newcommand{\mc}[1]{\ensuremath{\mathcal{#1}}}
\newcommand{\mb}[1]{\ensuremath{\mathbf{#1}}}
\newcommand{\mbb}[1]{\ensuremath{\mathbb{#1}}}

\usepackage{cvpr}
\cvprfinalcopy
\renewcommand{\subsection}[1]{\vspace{5pt} \noindent \textbf{#1:}}

\def\cvprPaperID{1572} 

\ifcvprfinal\fi

\begin{document}

\title{Bit-Planes: Dense Subpixel Alignment of Binary Descriptors}

\author{Hatem Alismail\\
Robotics Institute\\
Carnegie Mellon University\\
{\tt\small halismai@cs.cmu.edu}
\and
Brett Browning\\
Robotics Institute\\
Carnegie Mellon University\\
{\tt\small brettbrowning@gmail.com}
\and
Simon Lucey\\
Robotics Institute\\
Carnegie Mellon University\\
{\tt\small slucey@cs.cmu.edu}}

\maketitle

\begin{abstract}
Binary descriptors have been instrumental in the recent evolution of computationally efficient sparse image alignment algorithms. Increasingly, however, the vision community is interested in dense image alignment methods, which are more suitable for estimating correspondences from high frame rate cameras as they do not rely on exhaustive search. However, classic dense alignment approaches are sensitive to illumination change. In this paper, we propose an easy to implement and low complexity dense binary descriptor, which we refer to as bit-planes, that can be seamlessly integrated within a multi-channel Lucas \& Kanade framework. This novel approach combines the robustness of binary descriptors with the speed and accuracy of dense alignment methods. The approach is demonstrated on a template tracking problem achieving state-of-the-art robustness and faster than real-time performance on consumer laptops (400+ fps on a single core Intel i7) and hand-held mobile devices (100+ fps on an iPad Air 2).
\end{abstract}


\section{Introduction}
Binary descriptors such as BRIEF~\cite{brief} \& BRISK~\cite{brisk} are powerful tools for solving \emph{sparse} image alignment problems due to their discriminative power, robustness to illumination change, and low complexity~\cite{heinly2012,figat2014,Hafner2013,Vogel2013}. Matching binary descriptors is typically performed by exhaustive search~\cite{bostanci14,Kaneko20022223} using the Hamming distance. Exhaustive search, however, is inefficient when dense correspondences are required in real-time~\cite{grauman2013learning,muja2012fast}.

A classical way of speeding up the task of image alignment is to linearize pixel intensities of an image with respect to geometric displacement. The most notable example of this strategy can be found in the seminal work of Lucas \& Kanade~\cite{lk}. The Lucas \& Kanade (LK) algorithm attempts to establish an approximate linear relationship between appearance and geometric displacement. Efficient linear solvers can then be employed for finding the best geometric alignment of the image with respect to a known template. The relationship between appearance and geometric displacement is seldom linear, so the linearization process is typically repeated until  convergence. Due to its importance, numerous extensions and variations upon the LK algorithm have subsequently been explored in literature~\cite{baker2004lucas}.

At the heart of the LK algorithm is the notion that an approximate linear relationship between pixel appearance and geometric displacement can be reliably established. Pixel intensities are not deterministically differentiable with respect to geometric displacement. Instead, the linear relationship is established stochastically through spatial finite differences whose outputs we refer to as image gradients. The notion of estimating stochastic gradients on image intensities has a long and rich history dating back to some of the most seminal works of computer vision~\cite{Marr82}. Further, it has been well documented that pixel intensities within natural images are strongly correlated over small spatial areas further validating the assumed approximate linear relationship between pixel intensities and geometric displacement~\cite{simoncelli2001}. Pixel intensities, however, have a problem when applied to most practical image alignment tasks. Specifically, they violate the brightness constancy assumption, which states that pixel intensities describing a scene shall remain constant under geometric distortion. Our proposed dense bit-planes descriptor offers a solution to this shortcoming using a computationally efficient strategy.

\subsection{Contributions}
In this work we explore the validity of a~\emph{descriptor constancy} assumption using photometrically invariant descriptors. In particular, we explore the effectiveness of one of the simplest and most efficient binary descriptors LBP~\cite{lbp} --- or the Census Transform~\cite{zabih1994non} --- for robust and efficient dense correspondence estimation problems. The concept of linearizing feature descriptors with respect to geometric displacement within the LK framework is a relatively new and emerging topic~\cite{bristow14,Antonakos15}. Hitherto, most of the previously employed descriptors have a considerable computational footprint such as HOG~\cite{hog}, dense SIFT~\cite{Bosch2007,siftflow}, and even SURF~\cite{surf} making them unsuitable for practical use in many vision applications requiring dense correspondences in real-time from high frame rate data. In this paper we make the following contributions:
\begin{compactitem}
\item We propose the bit-planes descriptor, an adaptation of the LBP descriptor, that can be used within the LK framework. Specifically, we propose a multi-channel LK adaptation that allows us to minimize the Hamming distance using standard least squares optimization.
\item The suitability of our bit-planes descriptor for linearization is explored as a function of geometric displacement. We demonstrate that even though the dense bit-planes descriptor is inherently discontinuous it shares the same critical properties enjoyed by pixel intensities, which make them suitable for gradient-based optimization.
\item Unlike classical dense descriptors such as HOG and dense SIFT, we demonstrate the efficiency of our bit-planes descriptor on planar target tracking achieving speeds in excess of 400 fps on a laptop, and in excess of 120 fps on mobile devices. Furthermore, we demonstrate faster and more robust template tracking in comparison to RANSAC-based algorithms on sparse features, especially with low- and ambiguously textured objects.
\end{compactitem}

\section{The Lucas \& Kanade Algorithm}
In this section we briefly review the LK algorithm in order to introduce notation. Let $\mb{I}_0 : \mbb{R}^2 \rightarrow \mbb{R}$ be the template/reference image. After camera motion with parameter vector $\bm{\theta}\in\mbb{R}^p$, we obtain an input/moving image $\mb{I}_1$. We desire to estimate the parameters of motion such that the following quantity is minimized:
\begin{align}\label{eq:lk}
  \mc{E}(\mb{x}; \bm{\theta}) = \sum_{\mb{x} \in \Omega_0} \lVert \mb{I}_0(\mb{x}) -
  \mb{I}_1(\mb{x}'(\bm{\theta})) \rVert^2_2,
\end{align}
where $\Omega_0$ is a subset of pixels in the template image, $\bm{\theta}$ is an initial estimate of parameters and $\mb{x}'(\bm{\theta})$ describes the transformed pixel coordinates given the motion parameters, commonly known as the \emph{warping} function. By performing a first-order Taylor expansion of \cref{eq:lk} in the vicinity of $\bm{\theta}$, taking the derivative with respect to the parameter update, and equating it to zero, we arrive at the normal equations:
\begin{align}
  \mb{J}(\mb{x}; \bm{\theta})\T \mb{J}(\mb{x}; \bm{\theta})
  \Delta\bm{\theta} = \mb{J}(\mb{x}; \bm{\theta})\T \mb{e}(\mb{x};\bm{\theta}),
\end{align}
where $\mb{J}(\mb{x}; \bm{\theta})$ is the matrix of partial derivatives of the warped image intensities with respect to the motion parameters evaluated at the current estimate of
parameters $\bm{\theta}$, and $\mb{e}(\mb{x};\bm{\theta}) = \mb{I}_0(\mb{x}) - \mb{I}_1(\mb{x}'(\bm{\theta}))$. Using the chain rule we obtain
\begin{align}
  \mb{J}(\mb{x}; \bm{\theta}) = \frac{\partial\mb{I}_1(\mb{x})}{\partial\bm{\theta}} =
  \frac{\partial\mb{I}}{\partial\mb{x}'}
  \frac{\partial\mb{x}'}{\partial\bm{\theta}}.
\end{align}
where~$\nicefrac{\partial\mb{I}_{1}}{\partial\mb{x}'}$ is estimated stochastically through $x$- and~$y$- finite difference filters, while $\nicefrac{\partial\mb{x}'}{\partial\bm{\theta}}$ is obtained deterministically using the closed-form of the warping function. The original formulation of LK is applicable to a wide variety of problems. For special warps that satisfy a group requirement, however, a more efficient variation is Baker \& Matthews' Inverse Compositional algorithm (IC)~\cite{baker2004lucas} which we will use in the experimental portion of this paper.

\subsection{Photometric variation}
The classical formulation of LK relies on the brightness constancy assumption~\cite{lk}, which is seldom satisfied in real life applications. Techniques to address illumination change include: (i) estimating illumination parameters alongside the motion parameters (either jointly~\cite{bartoli2008groupwise} or in an alternating fashion~\cite{Tzimiropoulos2013}), (ii) using intrinsically robust similarity measures, such as Mutual Information~\cite{Dowson2008,Dame2010}, or the normalized correlation~\cite{Irani98}, and (iii) preprocessing the images to obtain a representation that is more robust to intensity variations~\cite{Lucey2013,Antonakos15,Wedel2009}.

On the one hand, estimating illumination is sensitive to the modeling assumptions and increases the dimensionality of the state and vector, thereby increasing the complexity of the optimization. On the other hand, optimizing robust similarity metrics requires general purpose optimizers that cannot exploit the special structure of least squares problems.

Preprocessing the image does not typically require restrictive assumptions, and does not affect the dimensionality of the state vector. Traditionally, preprocessing an image is done by convolving with filters, or other simple operations such as whitening the signal~\cite{Fleet2006,Sun2010}. Densely sampled feature descriptors are another form of preprocessing, which we adopt in this work. In particular, we propose the use of a dense bit-planes descriptor. During evaluation, we show that our approach exceeds the robustness of algorithms that explicitly model illumination as well as methods that rely on robust cost metrics. Furthermore, our method is more efficient, and simpler to implement. Central to our work is the multi-channel formulation of LK, which we review next.

\subsection{Multi-channel LK}
\begin{figure}
  \centering\small
  \includegraphics[width=\linewidth]{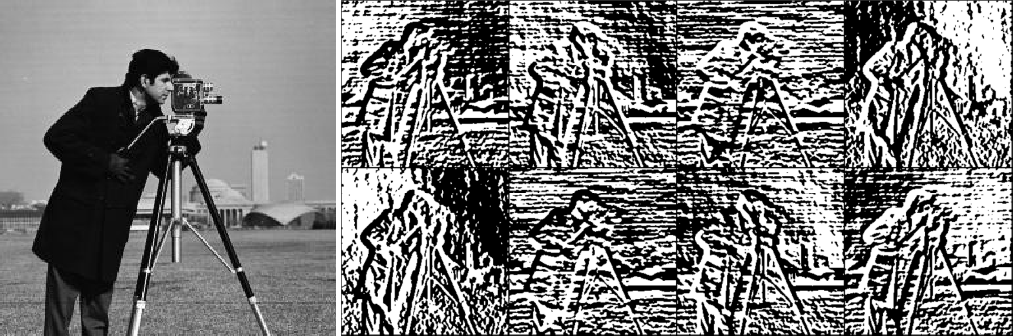}
  \caption{An example of the LBP descriptor evaluated on a $3\times 3$ neighborhood, which results in a $8-$channel bit-plane descriptor. Unlike the classic LBP descriptor, the bit-plane descriptor can be employed within a multi-channel LK framework using a sum of squared differences (SSD) cost measure.}\label{fig:channels}
\end{figure}

In this section we present a generalization of the LK algorithm to accommodate the application of multi-channel descriptors. Herein, we shall refer to this generalization as the multi-channel LK algorithm. Let $\bm{\phi}_{0} : \mbb{R}^2 \rightarrow \mbb{R}^d$ be the~$d$-channel representation of the template/reference image. Employing a similar notation to the classical LK algorithm, after camera motion with parameter vector $\bm{\theta}\in\mbb{R}^p$, we obtain an input/moving $d$-channel representation $\bm{\phi}_1$. To align descriptors using LK we seek to minimize
\begin{align}\label{eq:phi}
  \mc{E}_{\bm{\phi}}(\mb{x}; \bm{\theta}) = \sum_{\mb{x} \in \Omega_0} \lVert
    \bm{\phi}_{0}(\mb{x}) -
    \bm{\phi}_{1}(\mb{x}'(\bm{\theta}))\rVert^2.
\end{align}
To linearize \cref{eq:phi} we must obtain an estimate of the Jacobian $\mb{J}_{\bm{\phi}}(\mb{x};\bm{\theta}) = \nicefrac{\partial\bm{\phi}}{\partial {\bm{\theta}}} \in \mbb{R}^{d \times p}$. Let the value of the $j$-th channel, as illustrated in \cref{fig:channels}, of the multi-channel representation be described as $\bm{\phi}^j(\mb{x})$, where $\bm{\phi}(\mb{x}) = [\bm{\phi}^{1}(\mb{x}), \ldots, \bm{\phi}^{d}(\mb{x})]\T$. The sought Jacobian for each channel in~\cref{eq:phi} can be obtained using the chain rule
\begin{align}
  \frac{\partial\bm{\phi}^{j}_1(\mb{x})}{\partial\bm{\theta}} =
  \frac{\partial\bm{\phi}^{j}_{1}}{\partial\mb{x}'}
  \frac{\partial\mb{x}'}{\partial\bm{\theta}}
\end{align}
for $j=1,\ldots,d$ where $\nicefrac{\partial\bm{\phi}^{j}_{1}}{\partial\mb{x}'}$ is estimated stochastically through $x$- and~$y$- finite difference filters on~$\bm{\phi}^{j}_{1}$, and $\nicefrac{\partial\mb{x}'}{\partial\bm{\theta}}$ is obtained deterministically from the warp function. The multi-channel~$d \times p$ Jacobian matrix can then be formed as
\begin{equation}
\mb{J}_{\bm{\phi}}(\mb{x}; \bm{\theta}) =
\frac{\partial\bm{\phi}_1(\mb{x})}{\partial\bm{\theta}} =
\begin{bmatrix}
 \nicefrac{\partial\bm{\phi}^{1}_1(\mb{x})}{\partial\bm{\theta}} \\
 \vdots \\
 \nicefrac{\partial\bm{\phi}^{d}_1(\mb{x})}{\partial\bm{\theta}}
\end{bmatrix}\;.
\end{equation}
Using this multi-channel linearization all extensions and variations of the LK algorithm can be extended to different types of multi-channel descriptors. Recent work has demonstrated the utility of multi-channel LK~\cite{Antonakos15,bristow14} using classical dense descriptors such as dense SIFT and HOG\@. A novel component of this paper is the derivation of a low-complexity dense binary descriptor that can be seamlessly applied within the multi-channel LK framework.

\section{Dense Binary Descriptors}
Local Binary Patterns (LBP)~\cite{lbp} were among the first binary descriptors proposed in vision. An almost identical binary representation was independently developed by Zabih \& Woodfill under the name: Census Transform (CT)~\cite{zabih1994non}, which is still commonly applied in stereo and optical flow research~\cite{Vogel2013,Hafner2013,Muller2011,Stein2004}. LBP is based on the predicate of pixel comparisons in a small neighborhood as illustrated in \cref{fig:census}. By definition, the LBP descriptor is invariant to monotonic illumination change, which is desirable in practical image alignment applications. Recently, binary descriptor research has progressed significantly with the development of several high performance descriptors such as ORB~\cite{brief} and BRISK~\cite{brisk} among others~\cite{brief,binboost,bold,latch}.

\subsection{LBP descriptor}
When extracting a LBP descriptor about a pixel position~$\mb{x}$ one obtains,
\begin{align}
  \bm{\phi}(\mb{x}) = \sum_{i=1}^8 2^{i-1}
  \big[ \mb{I}(\mb{x}) \bowtie \mb{I}(\mb{x} + \Delta \mb{x}_{i})\big],
\label{Eq:LBP}
\end{align}
where~${\{\Delta \mb{x}_{i}\}}_{i=1}^{8}$ is the set of the eight relative coordinate displacements possible within a $3 \times 3$ neighborhood around the center pixel location~$\mb{x}$. Other neighborhood sizes and sampling locations can be used, but we found a $3\times 3$ region to perform best. The operator $\bowtie\in\{>,\ge,<,\le\}$ is a pixel comparison/binary test, and the bracket notation represents the indicator function. We refer to the LBP descriptor described in Equation~\cref{Eq:LBP} as single-channel since its output is a scalar at every pixel position~$\mb{x}$ within the image. A visual depiction of the single-channel LBP descriptor estimation process can be found in~\cref{fig:census}.
\begin{figure}
  \centering
  \begin{subfigure}[b]{0.25\linewidth}
  \centering
  \resizebox{\textwidth}{!}{\begin{tikzpicture}
      \fill[black!20!white] (0,0) rectangle (1,1);
      \draw[step=1cm,color=black!40!white] (-1,-1) grid (2,2);
      \node at (-0.5,+1.5) {\texttt{8}};
      \node at (+0.5,+1.5) {\texttt{12}};
      \node at (+1.5,+1.5) {\texttt{200}};

      \node at (-0.5,+0.5) {\texttt{56}};
      \node at (+0.5,+0.5) {\texttt{42}};
      \node at (+1.5,+0.5) {\texttt{55}};

      \node at (-0.5,-0.5) {\texttt{128}};
      \node at (+0.5,-0.5) {\texttt{16}};
      \node at (+1.5,-0.5) {\texttt{11}};
  \end{tikzpicture}}
  \caption{}\label{fig:ct1}
\end{subfigure}
\begin{subfigure}[b]{0.25\linewidth}
  \centering
  \resizebox{\textwidth}{!}{\begin{tikzpicture}
      \fill[black!20!white] (0,0) rectangle (1,1);
      \draw[step=1cm,color=black!40!white] (-1,-1) grid (2,2);
      \node at (-0.5,+1.5) {\tiny\texttt{8<42}};
      \node at (+0.5,+1.5) {\tiny\texttt{12<42}};
      \node at (+1.5,+1.5) {\tiny\texttt{200<42}};

      \node at (-0.5,+0.5) {\tiny\texttt{56<42}};
      \node at (+0.5,+0.5) {};
      \node at (+1.5,+0.5) {\tiny\texttt{55<42}};

      \node at (-0.5,-0.5) {\tiny\texttt{128<42}};
      \node at (+0.5,-0.5) {\tiny\texttt{16<42}};
      \node at (+1.5,-0.5) {\tiny\texttt{11<42}};
  \end{tikzpicture}}
  \caption{}\label{fig:ct2}
\end{subfigure}
\begin{subfigure}[b]{0.25\linewidth}
  \centering
  \resizebox{\textwidth}{!}{\begin{tikzpicture}
      \fill[black!20!white] (0,0) rectangle (1,1);
      \draw[step=1cm,color=black!40!white] (-1,-1) grid (2,2);
      \node at (-0.5,+1.5) {\texttt{1}};
      \node at (+0.5,+1.5) {\texttt{1}};
      \node at (+1.5,+1.5) {\texttt{0}};

      \node at (-0.5,+0.5) {\texttt{0}};
      \node at (+0.5,+0.5) {};
      \node at (+1.5,+0.5) {\texttt{0}};

      \node at (-0.5,-0.5) {\texttt{0}};
      \node at (+0.5,-0.5) {\texttt{1}};
      \node at (+1.5,-0.5) {\texttt{1}};
  \end{tikzpicture}}
  \caption{}\label{fig:ct3}
\end{subfigure}
  \caption{The canonical LBP descriptor is obtained by performing pixel comparisons in a fixed order and converting the binary string
  to a decimal value. In \cref{fig:ct1} the center pixel is in a $3\times 3$ neighborhood is highlighted, and compared to its neighbors as shown in \cref{fig:ct2}. Finally, the descriptor is obtained by combining the results of each comparison in \cref{fig:ct3} into a single scalar descriptor.}\label{fig:census}
\end{figure}

\subsection{Bit-planes descriptor}
When matching LBP descriptors it is common practice to employ the Hamming distance. Hamming distance is useful, because it matches LBP descriptors in a fashion that is invariant to the ordering of pixel comparisons within the~$3 \times 3$ neighborhood. Other distance metrics such as sum or squared distances (SSD) lack this desirable property and are dependent on the ordering specified by ${\{ \Delta \mb{x}_{i} \}}_{i=1}^{8}$. This becomes problematic when employing dense binary descriptors within the multi-channel LK framework due to its inherent dependence on the SSD\@.

To make dense binary descriptors compatible with LK we propose the~\emph{bit-planes} descriptor given by
\begin{align}
  \bm{\phi}(\mb{x}) = \begin{bmatrix} \mb{I}(\mb{x}) \bowtie \mb{I}(\mb{x} + \Delta \mb{x}_{1}) \\
\vdots \\ \mb{I}(\mb{x}) \bowtie \mb{I}(\mb{x} + \Delta \mb{x}_{8})
\end{bmatrix} \;.
\label{Eq:bitplanes}
\end{align}
For each pixel coordinate $\mb{x}$ in the image, this descriptor produces an $8$-channel vector of binary values~$\{0,1\}$. Notably, using the SSD with the multi-channel representation in~\cref{Eq:bitplanes} between two bit-planes descriptor is equivalent to the Hamming distance between single-channel LBP descriptors. Specifically, the ordering of the pixel comparisons within the~$3 \times 3$ neighborhood of the bit-planes descriptor has no effect on the SSD\@. An analysis on the choice of operator~$\bowtie\in\{>,\ge,<,\le\}$ is explored in the experiments section of this paper.

\section{Linearizing Bit-Planes}
In order for our proposed bit-planes descriptor to be effective within a multi-channel LK framework we first need to establish that there exists an approximate linear relationship between the multi-channel bit-planes descriptor and geometric displacements. Inspecting a visualization of the bit-planes descriptor in~\cref{fig:channels}, one could be doubtful about the existence of such relationship. Specifically, each channel of the bit-planes descriptor is highly discontinuous (due to its binary nature). In addition, estimating stochastic gradients per channel of the bit-planes descriptor seems strange as they can take on only three possibilities:~$\{-1, 0, +1\}$.

\subsection{SSD cost surface}
\begin{figure}
  \centering
  \begin{subfigure}{0.45\linewidth}
    \includegraphics[width=\linewidth]{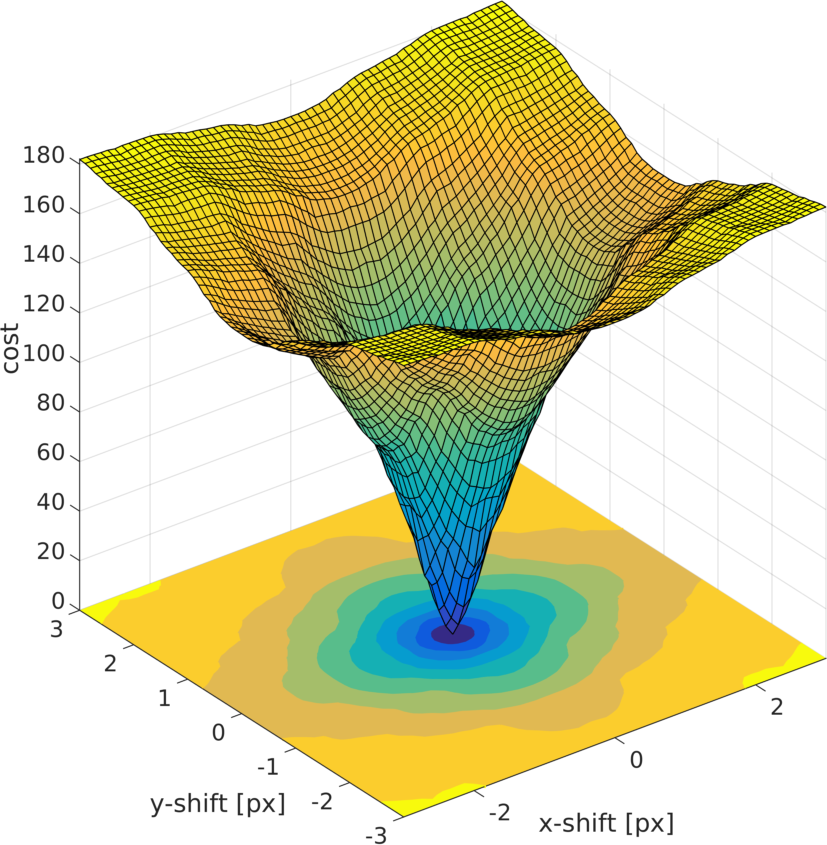}
    \caption{bit-planes}\label{fig:cost_bitplanes}
  \end{subfigure}
  \begin{subfigure}{0.45\linewidth}
    \includegraphics[width=\linewidth]{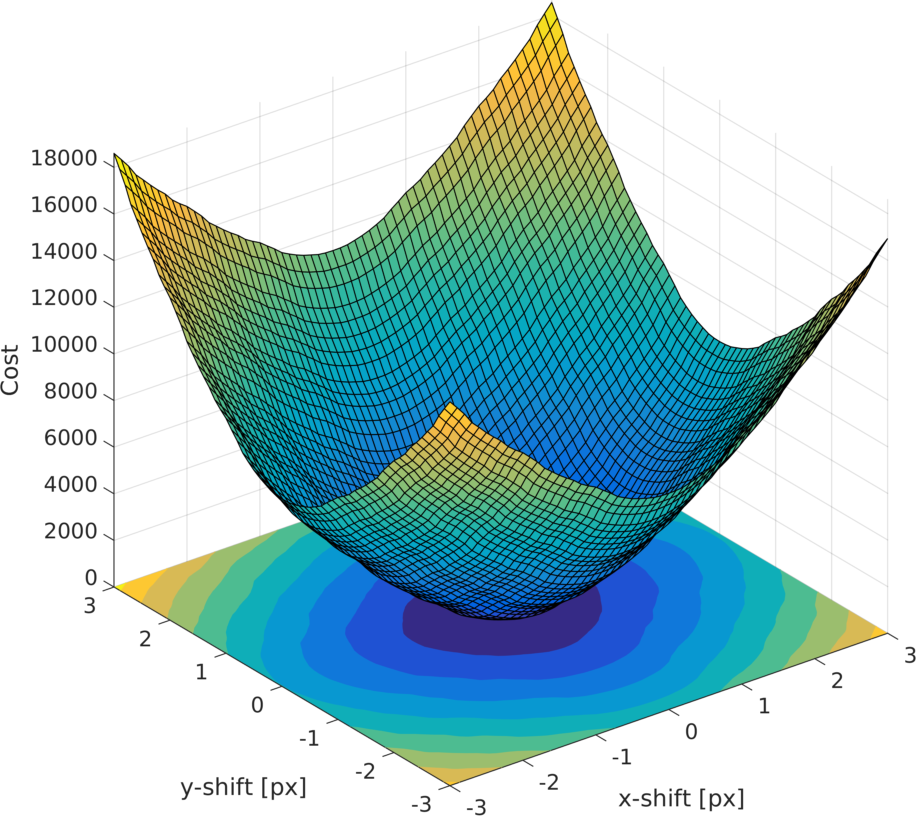}
    \caption{raw intensity}\label{fig:cost_intensity}
  \end{subfigure}
  \caption{Cost surface of our bit-planes descriptor \cref{fig:cost_bitplanes} computed over a subset of natural images~\cite{xiao2010sun} in comparison to SSD over raw intensity \cref{fig:cost_intensity}. Both cost surfaces are suitable for LK.}\label{fig:cost_surface}
\end{figure}

However, the news is not all gloomy. In~\cref{fig:cost_surface} we see the SSD cost surface between a patch within a natural image and shifted versions of itself in the $x$- and $y$- directions. This was repeated over a subset of natural images with the aggregate result being depicted in~\cref{fig:cost_surface}. Sub-pixel shifts are entertained here using bi-linear interpolation. In~\cref{fig:cost_intensity} one sees the cost surface for raw pixel intensities, and as expected, we see a quasi-convex cost surface surrounding the origin. This quasi-convex surface is important with respect to the effectiveness of the LK algorithm --- as the LK objective relies on a graceful reduction of the SSD cost as a function of geometric displacement from ground-truth. In fact the LK algorithm can be interpreted as attempting to hallucinate a convex quadratic representation of this SSD cost surface.
Interestingly, when inspecting~\cref{fig:cost_bitplanes} we see a similar quasi-convex cost surface, which indicates that bit-planes have similar properties to raw pixel intensities when it comes to the use of SSD as a measure of dissimilarity.

\subsection{Linear predictions of bit-planes}
\begin{figure}
  \small
  \centering
  \begin{subfigure}{0.45\linewidth}
    \includegraphics[width=\linewidth]{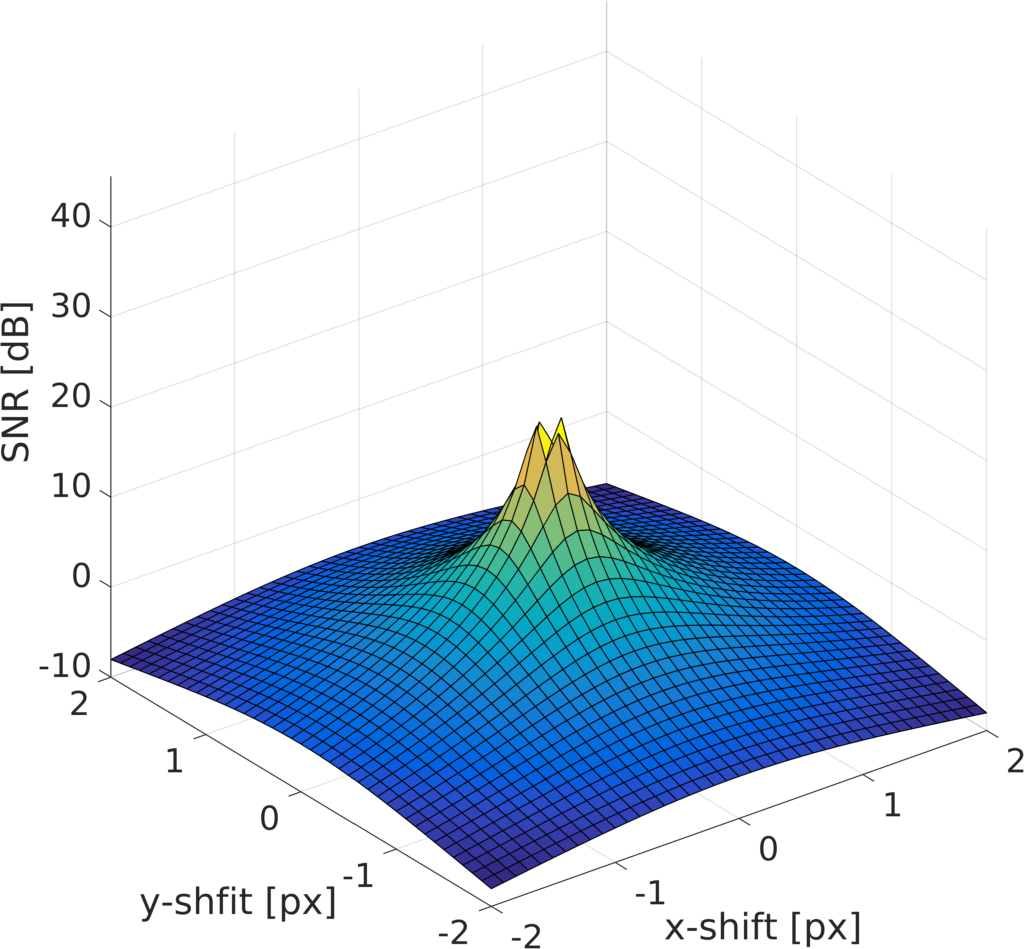}
    \caption{Bit-planes.}
  \end{subfigure}
  \begin{subfigure}{0.45\linewidth}
    \includegraphics[width=\linewidth]{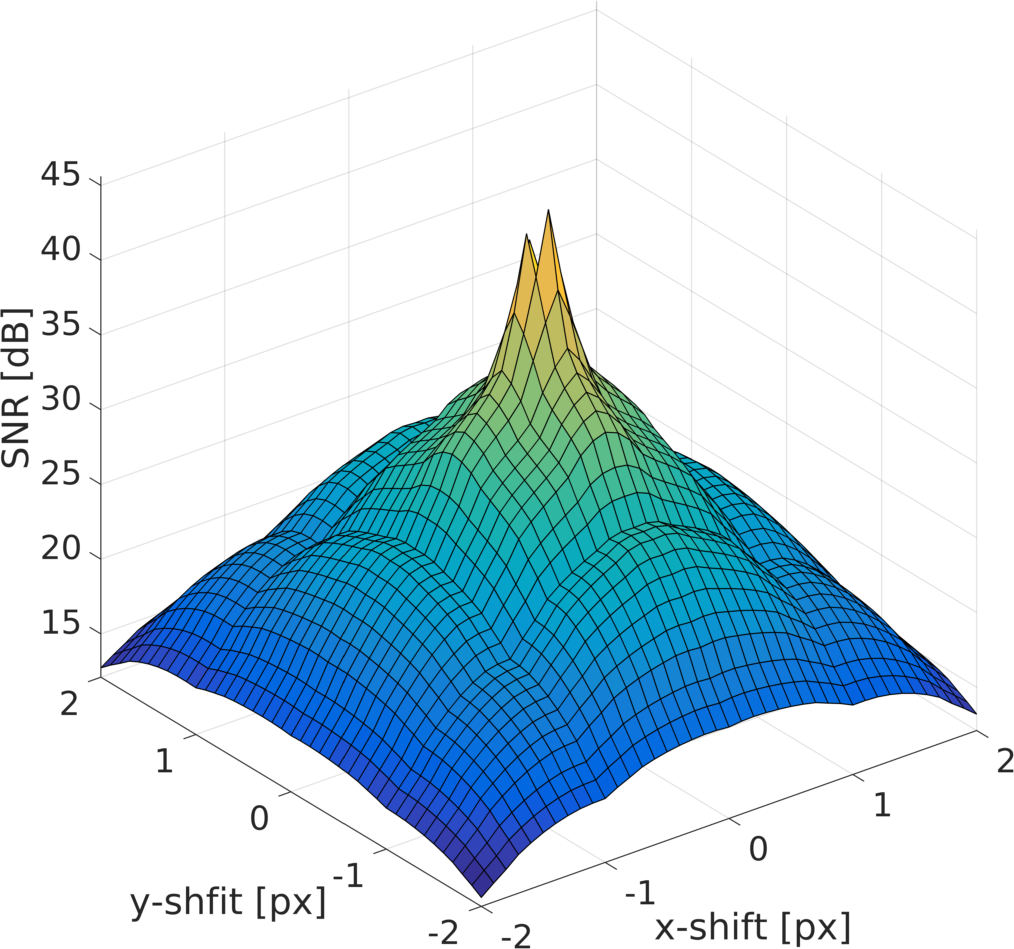}
    \caption{Raw Intensity.}
  \end{subfigure}
  \caption{Assessment of the linearization properties of our proposed bit-planes descriptor in terms of the signal-to-noise-ratio (SNR) as a function of translational displacement in~$x$- and~$y$- directions. One notes that even though raw pixels are superior, bit-planes offer a sufficient approximation to be used within a multi-channel LK framework. This thesis is the central focus of our paper.}\label{fig:snr}
\end{figure}

Consider a translational displacement warp~$\Delta \bm{\theta} \in \mbb{R}^{2}$ where we attempt to linearly predict an image representation~$\mb{R}$ (raw pixels, or bit-planes) in the~$x$- and~$y$- directions,
\begin{equation}
  \mb{R}(\mb{x}(\mb{0})) + \frac{\partial\mb{R}(\mb{0})}{\partial\bm{\theta}}\Delta \bm{\theta} \approx \mb{R}(\mb{x}(\Delta \bm{\theta}))\;\;.
\end{equation}
We employ the neutral notation~$\mb{R}$ to represent either raw pixels~$\mb{I}$ or bit-planes~$\bm{\phi}$. We can define the error of this linear approximation to be:
\begin{equation}
\bm{\epsilon}(\Delta \bm{\theta}) = \sum_{\mb{x} \in \Omega} \lVert \mb{R}(\mb{x}(\mb{0})) + \frac{\partial\mb{R}(\mb{0})}{\partial\bm{\theta}}\Delta \bm{\theta} - \mb{R}(\mb{x}(\Delta \bm{\theta})) \rVert^2_2
\end{equation}
The signal-to-noise-ratio (SNR) of the approximation can then be defined as
\begin{equation}
\mbox{SNR}(\Delta \bm{\theta}) = 10 \cdot [\log \sum_{\mb{x} \in \Omega} \lVert \mb{R}(\mb{x}(\mb{0})) \rVert^2_2  - \log \bm{\epsilon}(\Delta \bm{\theta})]
\end{equation}
In \cref{fig:snr} we depict the SNR of the linearized objective as a function of increasing translational shifts from the true minima for both raw intensities, and the binary channels in bit-planes. The experiments were carried out in a similar manner through the use of a subset of natural images and aggregated to form the results in \cref{fig:snr}. As expected, the SNR when using binary features is lower then using raw intensities. However, it seems that --- at least qualitatively --- bit-planes gradient estimates provide a good local linear approximation of the objective. Hence, further justifying the use of the bit-planes descriptor within the LK framework.

\section{Experiments}\label{sec:results}
In this section we shall attempt to answer a number of important questions regarding the validity of our dense bit-planes descriptor for robust and efficient image alignment.

\subsection{Pre-computing descriptors}
An obvious question to ask when considering the application of multi-channel descriptors, such as bit-planes, within the LK framework is: whether we can pre-compute the descriptors before warping? Specifically, due to the iterative nature of the LK algorithm it becomes computationally expensive to re-compute dense descriptors after each image warp. If one can pre-compute the descriptor before warping substantial efficiencies can be integrated into any LK-based image alignment.

We attempted to answer this question in \cref{fig:features_recompute} where we evaluated a number of well-known LK variants~\cite{baker2004lucas}: forward additive (FA), forward compositional (FC), and inverse compositional (IC) for the task of image alignment on natural images. Random warp initializations and appearance variation of the form
\begin{align}
  \mb{I}_1(\mb{x}) = 255 {\left(\frac{\alpha \mb{I}_0(\bm{\theta}_a(\mb{x})) +
  \beta}{255}\right)}^{1+\gamma}
\end{align}
were included, where $\bm{\theta}_a\left(\cdot\right)$ are the $6$DOF parameters of an affine warp, $\alpha$ and $\beta$ are multiplicative and additive lighting change terms, and $|\gamma| < 1$ is used for gamma correction. We can see that warping feature channels is less accurate than re-computing the descriptor on the warped image as shown in \cref{fig:features_recompute}.

\begin{figure}
  \centering
  \small
  \setlength\fwidth{0.8\linewidth}
  \setlength\fheight{0.25\linewidth}
%
%
\begin{tikzpicture}

\begin{axis}[%
width=0.951\fwidth,
height=\fheight,
at={(0\fwidth,0\fheight)},
scale only axis,
unbounded coords=jump,
xmin=0.5,
xmax=6.5,
xtick={1,2,3,4,5,6},
xticklabels={{FA},{FA-1},{FC},{FC-1},{IC},{IC-1}},
xmajorgrids,
ymin=0.00,
ymax=1.1,
ylabel={{\footnotesize Point RMS}},
ymajorgrids,
axis background/.style={fill=white}
]
\addplot [color=black,dashed,line width=1.0pt,forget plot]
  table[row sep=crcr]{%
1	0.325326627036463\\
1	0.67226693398939\\
};
\addplot [color=black,dashed,line width=1.0pt,forget plot]
  table[row sep=crcr]{%
2	0.319477135770505\\
2	0.637902247884905\\
};
\addplot [color=black,dashed,line width=1.0pt,forget plot]
  table[row sep=crcr]{%
3	0.269445578606686\\
3	0.533409746470148\\
};
\addplot [color=black,dashed,line width=1.0pt,forget plot]
  table[row sep=crcr]{%
4	0.277846426066235\\
4	0.563687683928354\\
};
\addplot [color=black,dashed,line width=1.0pt,forget plot]
  table[row sep=crcr]{%
5	0.906490527166485\\
5	1.59459468665002\\
};
\addplot [color=black,dashed,line width=1.0pt,forget plot]
  table[row sep=crcr]{%
6	0.906490527166479\\
6	1.59459468665002\\
};
\addplot [color=black,dashed,line width=1.0pt,forget plot]
  table[row sep=crcr]{%
1	0.0119391869424849\\
1	0.0914247795546039\\
};
\addplot [color=black,dashed,line width=1.0pt,forget plot]
  table[row sep=crcr]{%
2	0.00908676747570361\\
2	0.0910734456135793\\
};
\addplot [color=black,dashed,line width=1.0pt,forget plot]
  table[row sep=crcr]{%
3	0.00747820533830747\\
3	0.0895675265053522\\
};
\addplot [color=black,dashed,line width=1.0pt,forget plot]
  table[row sep=crcr]{%
4	0.00826092958855306\\
4	0.0867758719015193\\
};
\addplot [color=black,dashed,line width=1.0pt,forget plot]
  table[row sep=crcr]{%
5	0.00420338670968123\\
5	0.0419705432224481\\
};
\addplot [color=black,dashed,line width=1.0pt,forget plot]
  table[row sep=crcr]{%
6	0.00420338670968161\\
6	0.0419705432224479\\
};
\addplot [color=black,solid,line width=1.0pt,forget plot]
  table[row sep=crcr]{%
0.875	0.67226693398939\\
1.125	0.67226693398939\\
};
\addplot [color=black,solid,line width=1.0pt,forget plot]
  table[row sep=crcr]{%
1.875	0.637902247884905\\
2.125	0.637902247884905\\
};
\addplot [color=black,solid,line width=1.0pt,forget plot]
  table[row sep=crcr]{%
2.875	0.533409746470148\\
3.125	0.533409746470148\\
};
\addplot [color=black,solid,line width=1.0pt,forget plot]
  table[row sep=crcr]{%
3.875	0.563687683928354\\
4.125	0.563687683928354\\
};
\addplot [color=black,solid,line width=1.0pt,forget plot]
  table[row sep=crcr]{%
4.875	1.59459468665002\\
5.125	1.59459468665002\\
};
\addplot [color=black,solid,line width=1.0pt,forget plot]
  table[row sep=crcr]{%
5.875	1.59459468665002\\
6.125	1.59459468665002\\
};
\addplot [color=black,solid,line width=1.0pt,forget plot]
  table[row sep=crcr]{%
0.875	0.0119391869424849\\
1.125	0.0119391869424849\\
};
\addplot [color=black,solid,line width=1.0pt,forget plot]
  table[row sep=crcr]{%
1.875	0.00908676747570361\\
2.125	0.00908676747570361\\
};
\addplot [color=black,solid,line width=1.0pt,forget plot]
  table[row sep=crcr]{%
2.875	0.00747820533830747\\
3.125	0.00747820533830747\\
};
\addplot [color=black,solid,line width=1.0pt,forget plot]
  table[row sep=crcr]{%
3.875	0.00826092958855306\\
4.125	0.00826092958855306\\
};
\addplot [color=black,solid,line width=1.0pt,forget plot]
  table[row sep=crcr]{%
4.875	0.00420338670968123\\
5.125	0.00420338670968123\\
};
\addplot [color=black,solid,line width=1.0pt,forget plot]
  table[row sep=crcr]{%
5.875	0.00420338670968161\\
6.125	0.00420338670968161\\
};
\addplot [color=black,solid,line width=1.0pt,forget plot]
  table[row sep=crcr]{%
0.75	0.0914247795546039\\
0.75	0.325326627036463\\
1.25	0.325326627036463\\
1.25	0.0914247795546039\\
0.75	0.0914247795546039\\
};
\addplot [color=black,solid,line width=1.0pt,forget plot]
  table[row sep=crcr]{%
1.75	0.0910734456135793\\
1.75	0.319477135770505\\
2.25	0.319477135770505\\
2.25	0.0910734456135793\\
1.75	0.0910734456135793\\
};
\addplot [color=black,solid,line width=1.0pt,forget plot]
  table[row sep=crcr]{%
2.75	0.0895675265053522\\
2.75	0.269445578606686\\
3.25	0.269445578606686\\
3.25	0.0895675265053522\\
2.75	0.0895675265053522\\
};
\addplot [color=black,solid,line width=1.0pt,forget plot]
  table[row sep=crcr]{%
3.75	0.0867758719015193\\
3.75	0.277846426066235\\
4.25	0.277846426066235\\
4.25	0.0867758719015193\\
3.75	0.0867758719015193\\
};
\addplot [color=black,solid,line width=1.0pt,forget plot]
  table[row sep=crcr]{%
4.75	0.0419705432224481\\
4.75	0.906490527166485\\
5.25	0.906490527166485\\
5.25	0.0419705432224481\\
4.75	0.0419705432224481\\
};
\addplot [color=black,solid,line width=1.0pt,forget plot]
  table[row sep=crcr]{%
5.75	0.0419705432224479\\
5.75	0.906490527166479\\
6.25	0.906490527166479\\
6.25	0.0419705432224479\\
5.75	0.0419705432224479\\
};
\addplot [color=black,solid,line width=0.5pt,forget plot]
  table[row sep=crcr]{%
0.75	0.150180789208389\\
1.25	0.150180789208389\\
};
\addplot [color=black,solid,line width=0.5pt,forget plot]
  table[row sep=crcr]{%
1.75	0.151104911795183\\
2.25	0.151104911795183\\
};
\addplot [color=black,solid,line width=0.5pt,forget plot]
  table[row sep=crcr]{%
2.75	0.144538496103466\\
3.25	0.144538496103466\\
};
\addplot [color=black,solid,line width=0.5pt,forget plot]
  table[row sep=crcr]{%
3.75	0.143093632061048\\
4.25	0.143093632061048\\
};
\addplot [color=black,solid,line width=0.5pt,forget plot]
  table[row sep=crcr]{%
4.75	0.0744055289579334\\
5.25	0.0744055289579334\\
};
\addplot [color=black,solid,line width=0.5pt,forget plot]
  table[row sep=crcr]{%
5.75	0.0744055289579333\\
6.25	0.0744055289579333\\
};
\addplot [color=black,only marks,mark=+,mark options={mark size=1.618pt,solid,draw=black,line width=0.5pt},forget plot,line width=1pt]
  table[row sep=crcr]{%
nan	nan\\
};
\addplot [color=black,only marks,mark=+,mark options={mark size=1.618pt,solid,draw=black,line width=0.5pt},forget plot,line width=1pt]
  table[row sep=crcr]{%
2	0.749353159390059\\
};
\addplot [color=black,only marks,mark=+,mark options={mark size=1.618pt,solid,draw=black,line width=0.5pt},forget plot,line width=1pt]
  table[row sep=crcr]{%
3	0.542401152884196\\
3	0.667555874400955\\
};
\addplot [color=black,only marks,mark=+,mark options={mark size=1.618pt,solid,draw=black,line width=0.5pt},forget plot,line width=1pt]
  table[row sep=crcr]{%
4	0.673421934508278\\
};
\addplot [color=black,only marks,mark=+,mark options={mark size=1.618pt,solid,draw=black,line width=0.5pt},forget plot,line width=1pt]
  table[row sep=crcr]{%
5	3.36324935839486\\
};
\addplot [color=black,only marks,mark=+,mark options={mark size=1.618pt,solid,draw=black,line width=0.5pt},forget plot,line width=1pt]
  table[row sep=crcr]{%
6	3.36324935839485\\
};
\end{axis}
\end{tikzpicture}%
  \caption{Recomputing descriptors \emph{after} image warping is equivalent to warping feature channels given a linear relationship between the warp parameters and intensities (translation warps). The reason for reduced IC performance is due to higher noise at the template.}\label{fig:features_recompute_linear}
\end{figure}
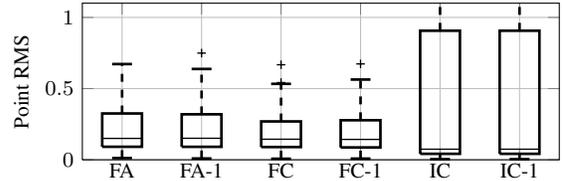

The degree to which warping the feature channels vs.~re-computing them affects accuracy depends on the application and the type of warp. For simple warps such as 2D translation, the relationship between intensity deformation as a function of warp parameters is linear. Hence, approximating multi-channel LK by warping the feature channels is equivalent to re-computing the features on warped images as shown in \cref{fig:features_recompute_linear}. However, for more complicated warps where deformation of image intensities is nonlinearly dependent on the warp parameters we expect a pronounced difference in alignment accuracy. This is because the value of each descriptor channel might significantly differ after a nonlinear warp. Overall, it is possible to approximate the multi-channel objective in \cref{eq:phi} with warping feature channels depending on the type and accuracy requirements of the application at hand. In our experiments, we chose to recompute descriptors after every iteration of image warping.

\begin{figure}
  \centering
  \small
  \setlength\fwidth{0.8\linewidth}
  \setlength\fheight{0.25\linewidth}
%
%
\begin{tikzpicture}

\begin{axis}[%
width=0.951\fwidth,
height=\fheight,
at={(0\fwidth,0\fheight)},
scale only axis,
xmin=0.5,
xmax=6.5,
xtick={1,2,3,4,5,6},
xticklabels={{FA},{FA-1},{FC},{FC-1},{IC},{IC-1}},
xmajorgrids,
ymin=0.003,
ymax=0.034,
ytick={0.003, 0.013, 0.024, 0.034},
ylabel={\footnotesize Point RMS},
ymajorgrids,
axis background/.style={fill=white}
]
\addplot [color=black,dashed,forget plot,line width=1pt]
  table[row sep=crcr]{%
1	0.0219965386492519\\
1	0.0267463805233228\\
};
\addplot [color=black,dashed,forget plot,line width=1pt]
  table[row sep=crcr]{%
2	0.016927533850461\\
2	0.0215901214882594\\
};
\addplot [color=black,dashed,forget plot,line width=1pt]
  table[row sep=crcr]{%
3	0.0219392720132254\\
3	0.0266905952340818\\
};
\addplot [color=black,dashed,forget plot,line width=1pt]
  table[row sep=crcr]{%
4	0.0171894031341911\\
4	0.0214073522841\\
};
\addplot [color=black,dashed,forget plot,line width=1pt]
  table[row sep=crcr]{%
5	0.0218301777680187\\
5	0.0272483709238291\\
};
\addplot [color=black,dashed,forget plot,line width=1pt]
  table[row sep=crcr]{%
6	0.0172426324557945\\
6	0.0201376979001312\\
};
\addplot [color=black,dashed,forget plot,line width=1pt]
  table[row sep=crcr]{%
1	0.0101549521263182\\
1	0.0167785662686705\\
};
\addplot [color=black,dashed,forget plot,line width=1pt]
  table[row sep=crcr]{%
2	0.00477795020616553\\
2	0.00906372711435472\\
};
\addplot [color=black,dashed,forget plot,line width=1pt]
  table[row sep=crcr]{%
3	0.00998399442991195\\
3	0.0167505134902565\\
};
\addplot [color=black,dashed,forget plot,line width=1pt]
  table[row sep=crcr]{%
4	0.004764155222485\\
4	0.00915619016271643\\
};
\addplot [color=black,dashed,forget plot,line width=1pt]
  table[row sep=crcr]{%
5	0.00987462756016741\\
5	0.0167748278249468\\
};
\addplot [color=black,dashed,forget plot,line width=1pt]
  table[row sep=crcr]{%
6	0.00484840061398891\\
6	0.00909445761151399\\
};
\addplot [color=black,solid,forget plot,line width=1pt]
  table[row sep=crcr]{%
0.875	0.0267463805233228\\
1.125	0.0267463805233228\\
};
\addplot [color=black,solid,forget plot,line width=1pt]
  table[row sep=crcr]{%
1.875	0.0215901214882594\\
2.125	0.0215901214882594\\
};
\addplot [color=black,solid,forget plot,line width=1pt]
  table[row sep=crcr]{%
2.875	0.0266905952340818\\
3.125	0.0266905952340818\\
};
\addplot [color=black,solid,forget plot,line width=1pt]
  table[row sep=crcr]{%
3.875	0.0214073522841\\
4.125	0.0214073522841\\
};
\addplot [color=black,solid,forget plot,line width=1pt]
  table[row sep=crcr]{%
4.875	0.0272483709238291\\
5.125	0.0272483709238291\\
};
\addplot [color=black,solid,forget plot,line width=1pt]
  table[row sep=crcr]{%
5.875	0.0201376979001312\\
6.125	0.0201376979001312\\
};
\addplot [color=black,solid,forget plot,line width=1pt]
  table[row sep=crcr]{%
0.875	0.0101549521263182\\
1.125	0.0101549521263182\\
};
\addplot [color=black,solid,forget plot,line width=1pt]
  table[row sep=crcr]{%
1.875	0.00477795020616553\\
2.125	0.00477795020616553\\
};
\addplot [color=black,solid,forget plot,line width=1pt]
  table[row sep=crcr]{%
2.875	0.00998399442991195\\
3.125	0.00998399442991195\\
};
\addplot [color=black,solid,forget plot,line width=1pt]
  table[row sep=crcr]{%
3.875	0.004764155222485\\
4.125	0.004764155222485\\
};
\addplot [color=black,solid,forget plot,line width=1pt]
  table[row sep=crcr]{%
4.875	0.00987462756016741\\
5.125	0.00987462756016741\\
};
\addplot [color=black,solid,forget plot,line width=1pt]
  table[row sep=crcr]{%
5.875	0.00484840061398891\\
6.125	0.00484840061398891\\
};
\addplot [color=black,solid,forget plot,line width=1pt]
  table[row sep=crcr]{%
0.75	0.0167785662686705\\
0.75	0.0219965386492519\\
1.25	0.0219965386492519\\
1.25	0.0167785662686705\\
0.75	0.0167785662686705\\
};
\addplot [color=black,solid,forget plot,line width=1pt]
  table[row sep=crcr]{%
1.75	0.00906372711435472\\
1.75	0.016927533850461\\
2.25	0.016927533850461\\
2.25	0.00906372711435472\\
1.75	0.00906372711435472\\
};
\addplot [color=black,solid,forget plot,line width=1pt]
  table[row sep=crcr]{%
2.75	0.0167505134902565\\
2.75	0.0219392720132254\\
3.25	0.0219392720132254\\
3.25	0.0167505134902565\\
2.75	0.0167505134902565\\
};
\addplot [color=black,solid,forget plot,line width=1pt]
  table[row sep=crcr]{%
3.75	0.00915619016271643\\
3.75	0.0171894031341911\\
4.25	0.0171894031341911\\
4.25	0.00915619016271643\\
3.75	0.00915619016271643\\
};
\addplot [color=black,solid,forget plot,line width=1pt]
  table[row sep=crcr]{%
4.75	0.0167748278249468\\
4.75	0.0218301777680187\\
5.25	0.0218301777680187\\
5.25	0.0167748278249468\\
4.75	0.0167748278249468\\
};
\addplot [color=black,solid,forget plot,line width=1pt]
  table[row sep=crcr]{%
5.75	0.00909445761151399\\
5.75	0.0172426324557945\\
6.25	0.0172426324557945\\
6.25	0.00909445761151399\\
5.75	0.00909445761151399\\
};
\addplot [color=black,solid,forget plot,line width=0.5pt]
  table[row sep=crcr]{%
0.75	0.0192789290561288\\
1.25	0.0192789290561288\\
};
\addplot [color=black,solid,forget plot,line width=0.5pt]
  table[row sep=crcr]{%
1.75	0.0124313460751006\\
2.25	0.0124313460751006\\
};
\addplot [color=black,solid,forget plot,line width=0.5pt]
  table[row sep=crcr]{%
2.75	0.0193351590014576\\
3.25	0.0193351590014576\\
};
\addplot [color=black,solid,forget plot,line width=0.5pt]
  table[row sep=crcr]{%
3.75	0.0123304803730193\\
4.25	0.0123304803730193\\
};
\addplot [color=black,solid,forget plot,line width=0.5pt]
  table[row sep=crcr]{%
4.75	0.0198846631329035\\
5.25	0.0198846631329035\\
};
\addplot [color=black,solid,forget plot,line width=0.5pt]
  table[row sep=crcr]{%
5.75	0.0121303766839266\\
6.25	0.0121303766839266\\
};
\addplot [color=black,only marks,mark=+,mark options={mark size=1.618pt,solid,draw=black,line width=0.5pt},forget plot,line width=1pt]
  table[row sep=crcr]{%
1	0.0298973063996007\\
1	0.030699767734481\\
1	0.031537226638695\\
};
\addplot [color=black,only marks,mark=+,mark options={mark size=1.618pt,solid,draw=black,line width=0.5pt},forget plot,line width=1pt]
  table[row sep=crcr]{%
2	0.0324658482181881\\
};
\addplot [color=black,only marks,mark=+,mark options={mark size=1.618pt,solid,draw=black,line width=0.5pt},forget plot,line width=1pt]
  table[row sep=crcr]{%
3	0.0298197133123936\\
3	0.0306239843196846\\
3	0.0314495936320871\\
};
\addplot [color=black,only marks,mark=+,mark options={mark size=1.618pt,solid,draw=black,line width=0.5pt},forget plot,line width=1pt]
  table[row sep=crcr]{%
4	0.0325730055171678\\
};
\addplot [color=black,only marks,mark=+,mark options={mark size=1.618pt,solid,draw=black,line width=0.5pt},forget plot,line width=1pt]
  table[row sep=crcr]{%
5	0.0303325900399528\\
5	0.03089606613773\\
5	0.0313123617947183\\
};
\addplot [color=black,only marks,mark=+,mark options={mark size=1.618pt,solid,draw=black,line width=0.5pt},forget plot,line width=1pt]
  table[row sep=crcr]{%
6	0.0327950716427288\\
};
\end{axis}
\end{tikzpicture}%
  \caption{Recomputing descriptors \emph{after} image warping shows consistently better performance than warping feature images when tested with several LK variants. FA: Forward Addition, FC: Forward Compositional, IC: Inverse Compositional. The suffix `-1` indicates \emph{recomputed} features on warped images.}\label{fig:features_recompute}
\end{figure}
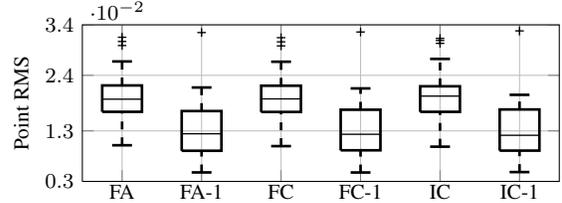

\subsection{LBP within LK}
Employing bit-planes requires the alignment of eight separate channels as opposed to a single channel when working with raw intensities. In \cref{sec:template_tracking} we discussed the problems of using a LBP descriptor within the LK framework. In particular, the representation is inherently sensitive to the ordering of pixel comparisons when using a SSD measure of dissimilarity. Using LBP descriptors within a LK framework as been reported to perform well~\cite{Vogel2013,Hafner2013} given small displacements. However, under moderate displacements the use of the LBP descriptor in LK introduces biases due to choices of the binary test and neighborhood ordering. In \cref{fig:census_bias} we show the effect of differing binary comparison operators $\bowtie\in\{>,\ge,<,\le\}$ compared to our proposed bit-plane descriptor. Our bit-planes descriptor is unaffected by the ordering. In our experiments we noticed indistinguishable differences in performance between binary comparison operators when employing the bit-planes descriptor. As a result, we chose to use the $>$ operator for the rest of our experiments.

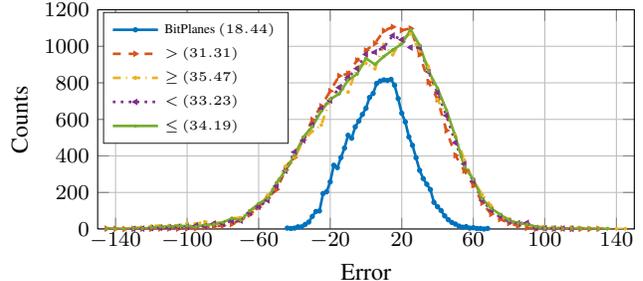
\begin{figure}
  \centering\small
  \setlength\fwidth{0.9\linewidth}
  \setlength\fheight{0.35\linewidth}
%
%
\definecolor{mycolor1}{rgb}{0.00000,0.44700,0.74100}%
\definecolor{mycolor2}{rgb}{0.85000,0.32500,0.09800}%
\definecolor{mycolor3}{rgb}{0.92900,0.69400,0.12500}%
\definecolor{mycolor4}{rgb}{0.49400,0.18400,0.55600}%
\definecolor{mycolor5}{rgb}{0.46600,0.67400,0.18800}%
\begin{tikzpicture}

\begin{axis}[%
width=0.951\fwidth,
height=\fheight,
at={(0\fwidth,0\fheight)},
scale only axis,
xmin=-150,
xmax=150,
xtick={-140,-100,-60,-20,20,60,100,140},
xlabel={Error},
xmajorgrids,
ymin=0,
ymax=1200,
ylabel={Counts},
ymajorgrids,
axis background/.style={fill=white},
legend style={legend cell align=left,align=left,draw=white!15!black,%
  at={(0.01,0.99)},anchor=north west},
y tick label style={/pgf/number format/.cd,%
          scaled y ticks = false,
          set thousands separator={},
          fixed}
]
\addplot [color=mycolor1,solid,line width=1.0pt,mark=*,mark options={solid,
  mark size=.5}]
  table[row sep=crcr]{%
-44	5\\
-42	3\\
-40	5\\
-38	12\\
-36	12\\
-34	25\\
-32	38\\
-30	69\\
-28	96\\
-26	99\\
-24	194\\
-22	204\\
-20	258\\
-18	349\\
-16	336\\
-14	391\\
-12	443\\
-10	513\\
-8	497\\
-6	559\\
-4	591\\
-2	627\\
0	661\\
2	720\\
4	742\\
6	797\\
8	810\\
10	816\\
12	813\\
14	819\\
16	787\\
18	714\\
20	634\\
22	586\\
24	504\\
26	458\\
28	390\\
30	321\\
32	276\\
34	245\\
36	216\\
38	146\\
40	122\\
42	101\\
44	72\\
46	51\\
48	36\\
50	27\\
52	16\\
54	10\\
56	13\\
58	5\\
60	7\\
62	3\\
64	3\\
66	0\\
68	3\\
};
\addlegendentry{{\tiny BitPlanes $(18.44)$}};

\addplot [color=mycolor2,dashed,line width=1.0pt,mark=triangle,mark options={solid,rotate=270,mark size=0.5pt}]
  table[row sep=crcr]{%
-145	2\\
-140	1\\
-135	0\\
-130	1\\
-125	2\\
-120	1\\
-115	3\\
-110	3\\
-105	2\\
-100	3\\
-95	5\\
-90	10\\
-85	14\\
-80	19\\
-75	27\\
-70	51\\
-65	79\\
-60	119\\
-55	170\\
-50	216\\
-45	322\\
-40	391\\
-35	487\\
-30	598\\
-25	672\\
-20	755\\
-15	838\\
-10	850\\
-5	923\\
0	977\\
5	1016\\
10	1085\\
15	1106\\
20	1088\\
25	1098\\
30	955\\
35	804\\
40	710\\
45	539\\
50	403\\
55	319\\
60	240\\
65	143\\
70	82\\
75	54\\
80	32\\
85	18\\
90	6\\
95	1\\
100	4\\
105	1\\
110	3\\
115	0\\
120	1\\
125	0\\
130	1\\
};
\addlegendentry{{\tiny $>$ $(31.31)$}};

\addplot [color=mycolor3,dashdotted,line width=1.0pt,mark=star,mark options={solid, mark size=0.5pt}]
  table[row sep=crcr]{%
-140	1\\
-135	2\\
-130	7\\
-125	9\\
-120	16\\
-115	19\\
-110	20\\
-105	20\\
-100	31\\
-95	29\\
-90	40\\
-85	34\\
-80	36\\
-75	59\\
-70	62\\
-65	101\\
-60	119\\
-55	184\\
-50	251\\
-45	343\\
-40	393\\
-35	484\\
-30	535\\
-25	569\\
-20	708\\
-15	780\\
-10	749\\
-5	885\\
0	908\\
5	877\\
10	1009\\
15	954\\
20	1038\\
25	1069\\
30	974\\
35	887\\
40	709\\
45	616\\
50	512\\
55	379\\
60	263\\
65	189\\
70	126\\
75	71\\
80	55\\
85	27\\
90	31\\
95	13\\
100	12\\
105	10\\
110	5\\
115	11\\
120	5\\
125	5\\
130	4\\
135	4\\
140	0\\
145	1\\
};
\addlegendentry{{\tiny $\ge$ $(35.47)$}};

\addplot [color=mycolor4,dotted,line width=1.0pt,mark=triangle,mark options={solid,rotate=90,mark size=0.5pt}]
  table[row sep=crcr]{%
-130	1\\
-125	2\\
-120	5\\
-115	7\\
-110	9\\
-105	11\\
-100	13\\
-95	19\\
-90	16\\
-85	30\\
-80	29\\
-75	44\\
-70	46\\
-65	88\\
-60	122\\
-55	157\\
-50	235\\
-45	319\\
-40	421\\
-35	485\\
-30	572\\
-25	660\\
-20	703\\
-15	774\\
-10	830\\
-5	878\\
0	956\\
5	968\\
10	995\\
15	1059\\
20	1036\\
25	993\\
30	993\\
35	860\\
40	746\\
45	621\\
50	472\\
55	356\\
60	262\\
65	157\\
70	103\\
75	71\\
80	43\\
85	32\\
90	34\\
95	6\\
100	4\\
105	3\\
110	3\\
115	1\\
};
\addlegendentry{{\tiny $<$ $(33.23)$}};

\addplot [color=mycolor5,solid,line width=1.0pt,mark=star,mark options={solid,mark size=0.5pt}]
  table[row sep=crcr]{%
-145	1\\
-140	2\\
-135	0\\
-130	3\\
-125	3\\
-120	7\\
-115	13\\
-110	14\\
-105	16\\
-100	17\\
-95	13\\
-90	23\\
-85	26\\
-80	36\\
-75	58\\
-70	48\\
-65	99\\
-60	139\\
-55	168\\
-50	252\\
-45	320\\
-40	403\\
-35	503\\
-30	560\\
-25	645\\
-20	710\\
-15	740\\
-10	815\\
-5	848\\
0	934\\
5	901\\
10	944\\
15	980\\
20	1013\\
25	1091\\
30	1005\\
35	864\\
40	770\\
45	651\\
50	505\\
55	341\\
60	272\\
65	166\\
70	113\\
75	80\\
80	48\\
85	32\\
90	26\\
95	13\\
100	4\\
105	5\\
110	2\\
115	2\\
120	3\\
125	2\\
130	0\\
135	1\\
};
\addlegendentry{{\tiny $\le$ $(34.19)$}};

\end{axis}
\end{tikzpicture}%
  \caption{Histogram of final intensity errors when using our proposed multi-channel bit-planes vs.~classical single-channel LBP descriptors with different comparison operators $\bowtie\in\{>,\ge,<,\le\}$, the RMS is shown in parenthesis.}\label{fig:bias_hist}
\end{figure}

\begin{figure}
  \centering\small
  \begin{subfigure}{0.49\linewidth}
    \includegraphics[width=\linewidth]{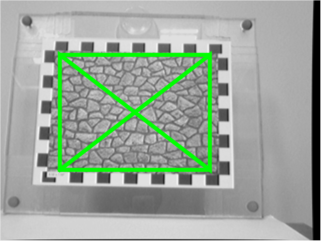}
    \caption{Template at $t=0$}
  \end{subfigure}
  \begin{subfigure}{0.49\linewidth}
    \includegraphics[width=\linewidth]{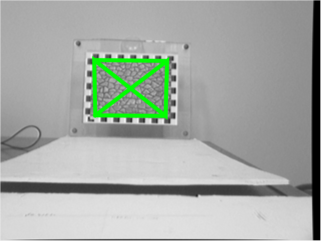}
    \caption{Bit-planes result at $t=50$}\label{fig:bitplanes_res1}
  \end{subfigure}
  \begin{subfigure}{0.24\linewidth}
    \includegraphics[width=\linewidth,trim={4cm 5.5cm 7cm 2.5cm},clip]{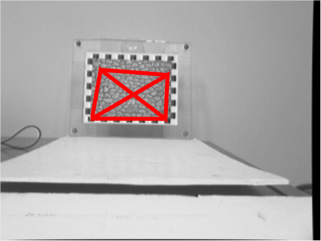}
    \caption{$\bowtie~\coloneqq~>$}
  \end{subfigure}
  \begin{subfigure}{0.24\linewidth}
    \includegraphics[width=\linewidth,trim={4cm 5.5cm 7cm 2.5cm},clip]{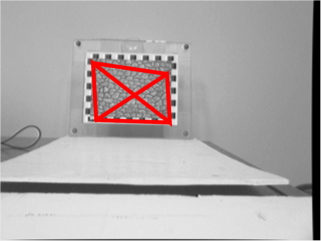}
    \caption{$\bowtie~\coloneqq~\ge$}
  \end{subfigure}
  \begin{subfigure}{0.24\linewidth}
    \includegraphics[width=\linewidth,trim={4cm 5.5cm 7cm 2.5cm},clip]{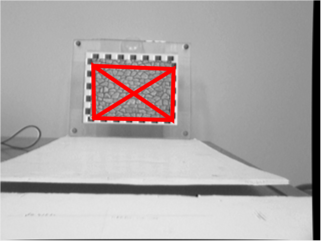}
    \caption{$\bowtie~\coloneqq~<$}
  \end{subfigure}
  \begin{subfigure}{0.24\linewidth}
    \includegraphics[width=\linewidth,trim={4cm 5.5cm 7cm 2.5cm},clip]{figs/ct_result_lt.png}
    \caption{$\bowtie~\coloneqq~\le$}
  \end{subfigure}
  \caption{Tracking drift when using LBP vs.~bit-planes. The bottom row shows the result of template tracking using LPB\@. The image magnified for better visualization (compare with \cref{fig:bitplanes_res1}). A quantitative analysis is shown in \cref{fig:bias_hist}. Best viewed in color.}\label{fig:census_bias}
\end{figure}

\subsection{Real-time template tracking}\label{sec:template_tracking}
We evaluate the performance of bit-planes for a template tracking problems using the benchmark dataset collected by Gauglitz \etal~\cite{Gauglitz_IJCV2011}. An example of the dataset is shown in \cref{fig:tracking_bricks}. Our plane tracker estimates an 8DOF homography using the IC algorithm~\cite{baker2004lucas}. The template is extracted from the first frame in each sequence and is kept fixed throughout the sequence as we are interested in tracking robustness overtime. To improve convergence we use a 3-level pyramid and initialize the tracker for subsequent frames using the most recent estimate. We use Gauss-Newton as the optimization algorithm, without robust weighting, with a maximum of 100 iterations. Tracking terminates early if the relative change in the estimated parameters between iterations drops below $1\times 10^{-6}$, or the relative change in the cost function reduction drops below $1\times 10^{-5}$. For small motion, the tracker typically converges in less than $10$ iterations. Our implementation runs faster than real time as shown in \cref{table:runtime}. The efficiency is achieved by utilizing SIMD instructions on the CPU that allows us to process $16$ pixels at once (or $32$ pixels with modern AVX instructions). Additionally, the operations required to compute the LBP descriptor are limited to bit shifts, ORs and ANDs, all of which can be performed with low latency. 

\begin{table}
  \centering\small
  \begin{tabular}{l c c}
    \toprule
    Algorithm & \# parameters & \# channels                     \\ \midrule
    \textbf{BP} (ours)                            & $8$   & $8$ \\
    \textbf{ECC}~\cite{evangelidis2008parametric} & $8$   & $1$ \\
    \textbf{DIC-1}~\cite{bartoli2008groupwise}    & $10$  & $1$ \\
    \textbf{DIC-2}~\cite{bartoli2008groupwise}    & $20$  & $3$ \\
    \textbf{DF}~\cite{crivellaro2014robust}       & $8$   & $5$ \\
    \textbf{GC}~\cite{Brox2004}                   & $8$   & $3$ \\
    \textbf{GM}                                   & $8$   & $2$ \\
    \textbf{LK}                                   & $8$   & $1$ \\
    \bottomrule
  \end{tabular}
  \caption{Algorithms compared in this work. The number of parameters indicates the DOF of the state vector, which is $8$ for a homography in addition to any photometric parameters. We use the authors' code for ECC and DIC.}\label{table:algs}
\end{table}

We compare the performance of our algorithm against a variety of template tracking methods summarized in \cref{table:algs}. The algorithms are: the enhance correlation coefficient \textbf{ECC}~\cite{evangelidis2008parametric}, which serves as an example of an intrinsically robust cost function that is invariant up to an affine illumination change. The Dual Inverse Compositional (DIC) algorithm~\cite{bartoli2008groupwise}, which severs as an example of algorithms that attempt to estimate illumination parameters. We use two variations of the DIC: (i) the gain+bias model on grayscale images denoted by \textbf{DIC-1}, and (ii) using a full affine lighting model the makes use of RGB image data denoted by \textbf{DIC-2}\@. We also compare the performance against a recently published descriptor-based method~\cite{crivellaro2014robust} called Descriptor Fields \textbf{DF}. Finally, we include baseline results from raw intensity \textbf{BF}, improved LK with the Gradient Constraint \textbf{GC}~\cite{Brox2004}, and alignment with the Gradient Magnitude \textbf{GM}\@.

We report two quantities in the evaluation. First, is the percentage of successfully tracked frames. A frame is successfully tracked if the overlap between the estimate and the ground truth is greater than $90\%$. The overlap is computed as $o = (A \cap B) / (A \cup B)$, where $A$ is the warped image given each algorithm's estimate, and $B$ is the warped image given the ground truth. Second, since we are also interested in subpixel accuracy we show the mean percentage of overlap across all frames given by $m = \nicefrac{1}{n}\sum_{i=1}^n o_i$, where $n$ is the number of frames in each sequence.

\subsection{Real-time results}
\begin{figure}
  \centering\small
  \begin{subfigure}{\linewidth}
    \centering
  \includegraphics[width=0.23\linewidth,trim={2cm 4cm 5cm 4cm},clip]{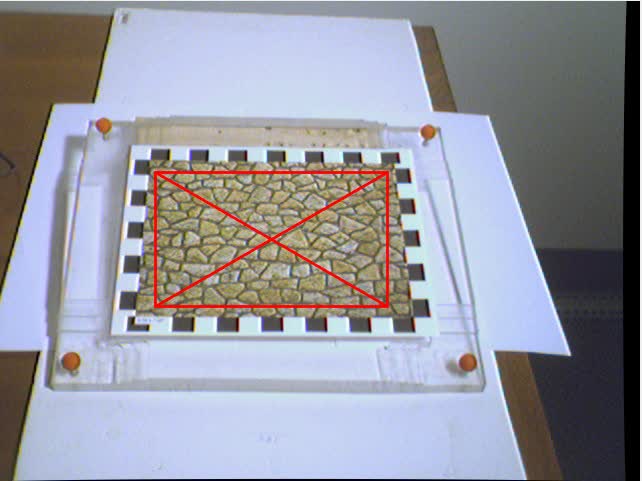}
  \includegraphics[width=0.23\linewidth,trim={2cm 4cm 5cm 4cm},clip]{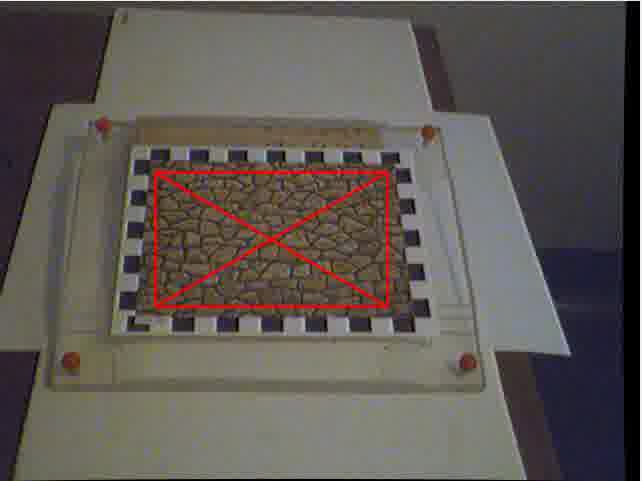}
  \includegraphics[width=0.23\linewidth,trim={2cm 4cm 5cm 4cm},clip]{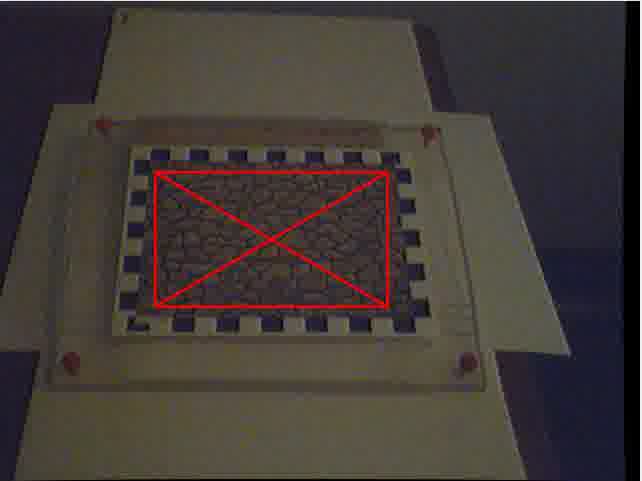}
  \includegraphics[width=0.23\linewidth,trim={2cm 4cm 5cm 4cm},clip]{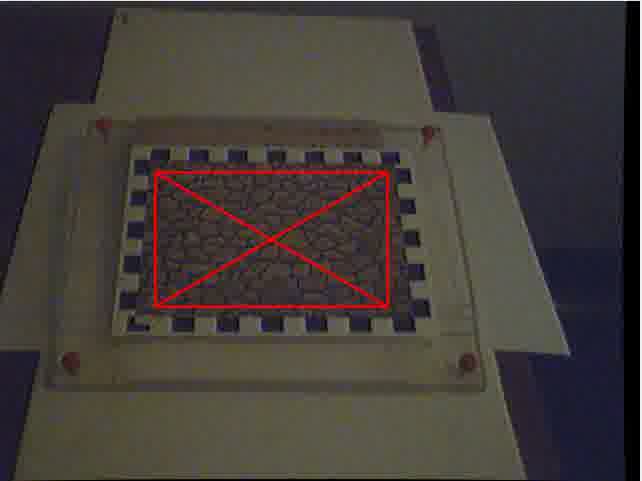}

  \includegraphics[width=0.23\linewidth,trim={2cm 4cm 5cm 4cm},clip]{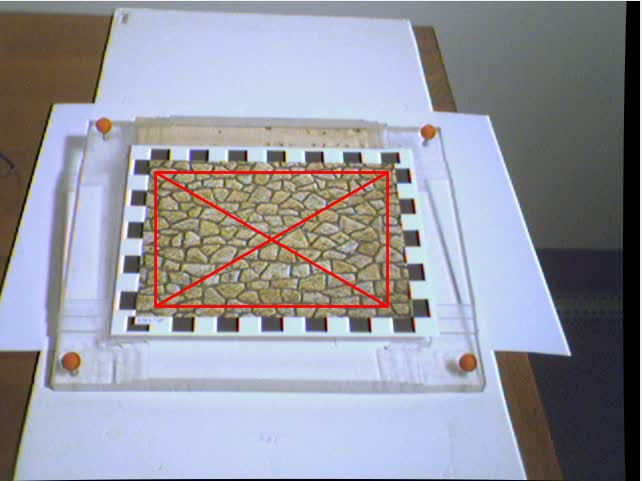}
  \includegraphics[width=0.23\linewidth,trim={2cm 4cm 5cm 4cm},clip]{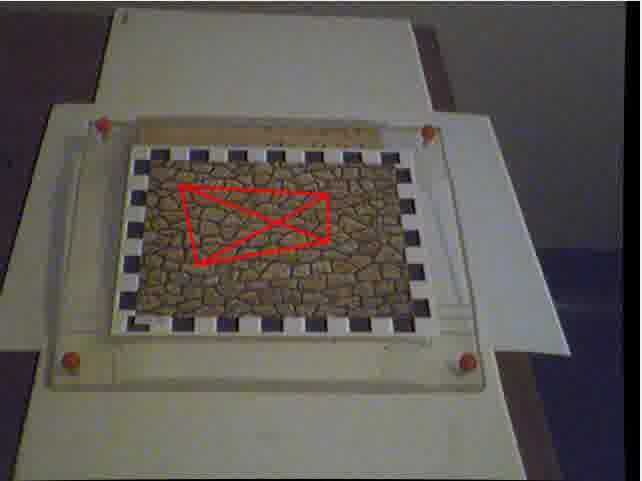}
  \includegraphics[width=0.23\linewidth,trim={2cm 4cm 5cm 4cm},clip]{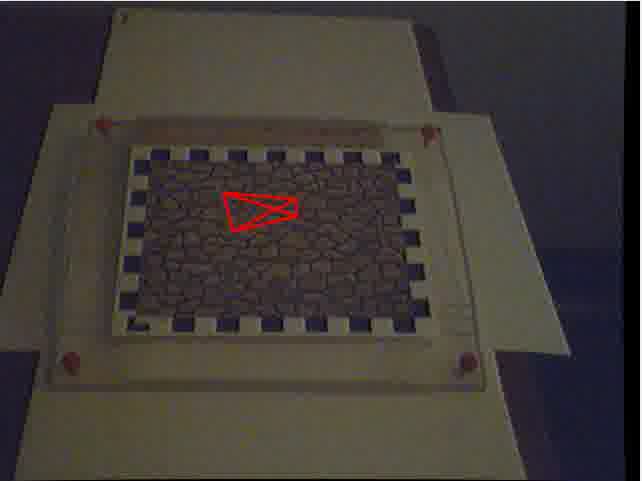}
  \includegraphics[width=0.23\linewidth,trim={2cm 4cm 5cm 4cm},clip]{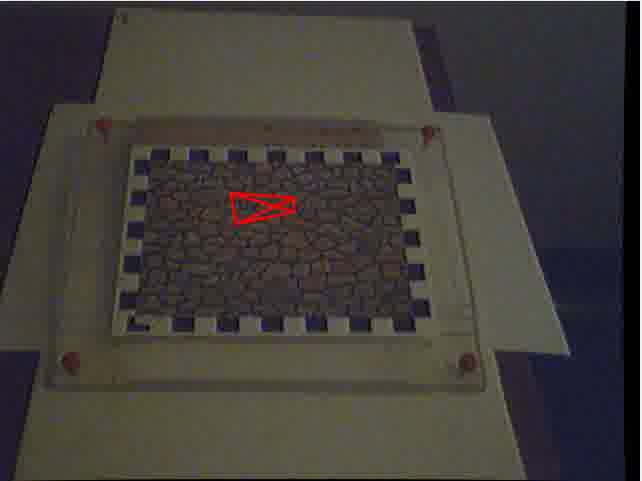}
  \caption{Lighting change.}
\end{subfigure}

\begin{subfigure}{\linewidth}
  \centering
  \includegraphics[width=0.23\linewidth,trim={1cm 1cm 5cm 3cm},clip]{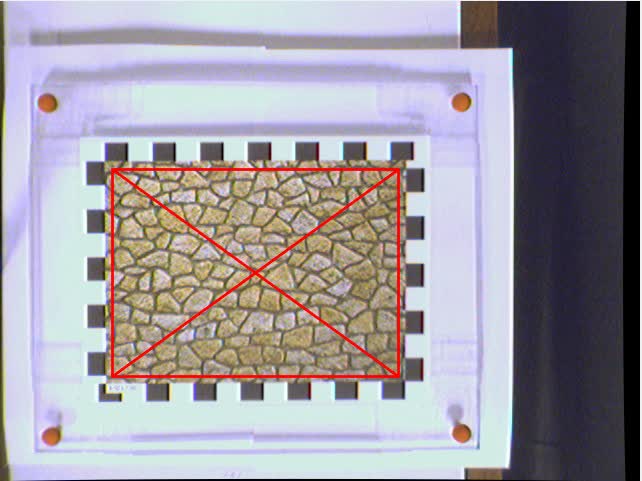}
  \includegraphics[width=0.23\linewidth,trim={1cm 1cm 5cm 3cm},clip]{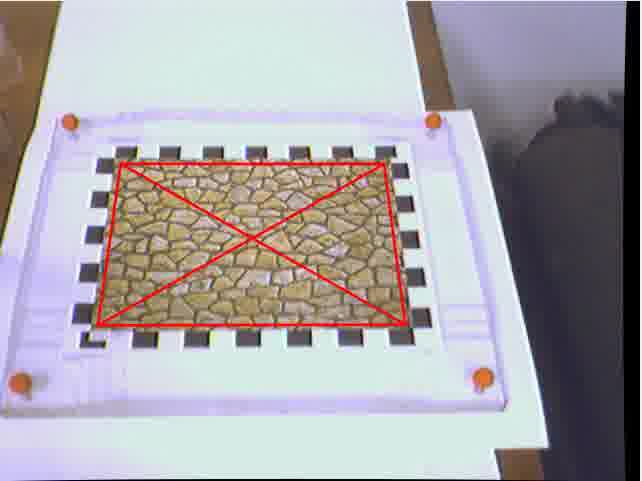}
  \includegraphics[width=0.23\linewidth,trim={1cm 1cm 5cm 3cm},clip]{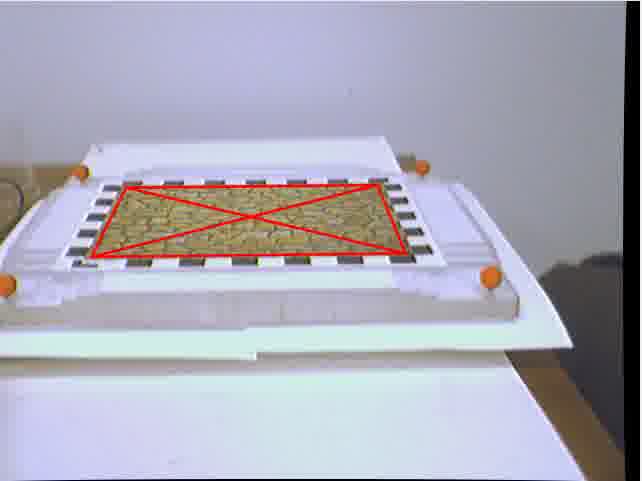}
  \includegraphics[width=0.23\linewidth,trim={1cm 1cm 5cm 3cm},clip]{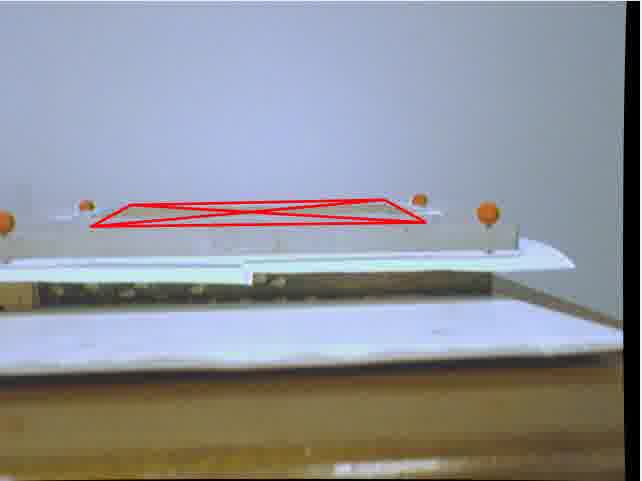}

  \includegraphics[width=0.23\linewidth,trim={1cm 1cm 5cm 3cm},clip]{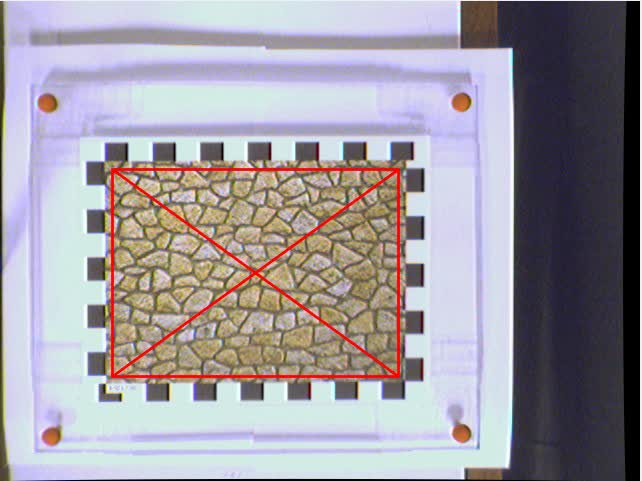}
  \includegraphics[width=0.23\linewidth,trim={1cm 1cm 5cm 3cm},clip]{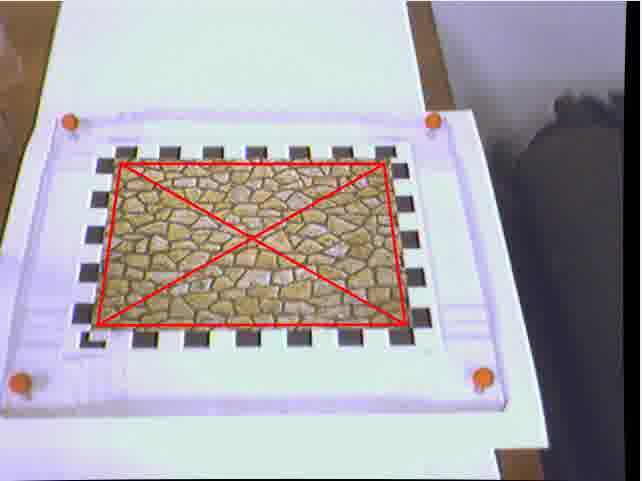}
  \includegraphics[width=0.23\linewidth,trim={1cm 1cm 5cm 3cm},clip]{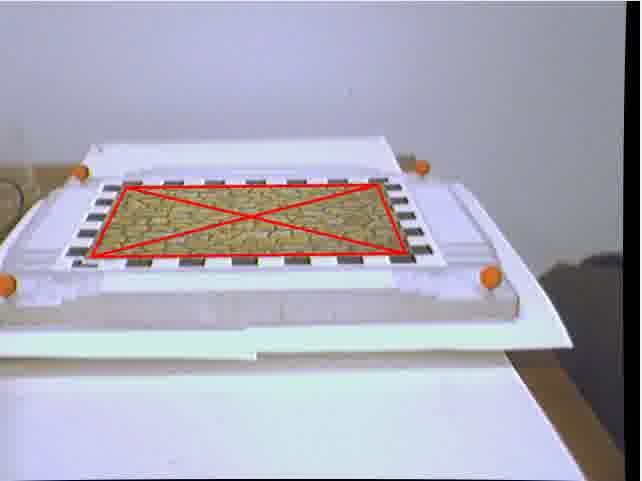}
  \includegraphics[width=0.23\linewidth,trim={1cm 1cm 5cm 3cm},clip]{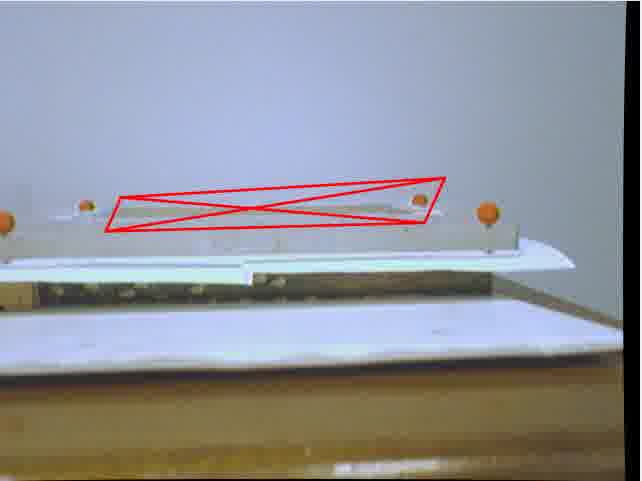}
  \caption{Out-of-plane rotation.}\label{fig:out_of_plane}
\end{subfigure}
\caption{Tracking results using the Bricks dataset~\cite{Gauglitz_IJCV2011}. The top row of each figure shows the performance of bit-planes, while the bottom row shows classical intensity-based LK\@.}\label{fig:tracking_bricks}
\end{figure}

Results are compared for three types of geometric and photometric variations. First is an \textbf{out of plane rotation}, which induces perspective change as shown in \cref{fig:out_of_plane}. Second, is \textbf{dynamic lighting change} where the image is stationary but a illuminated with nonlinearly varying light source. Finally a \textbf{static lighting change}, where the transition phase of change in lighting is omitted.

\begin{figure}
  \centering\small
  \setlength\fwidth{0.8\linewidth}
  \setlength\fheight{0.4\linewidth}
  \input{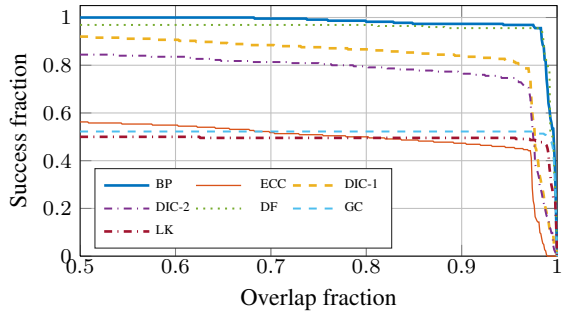}
  \caption{Fraction of successfully tracked frames as function of the overlap area given the ground truth. Bit-planes and DF perform better than other methods. However, in \cref{table:tracking_results} we see that bit-planes' performance is better with challenging sequences.}
\label{fig:auc}
\end{figure}

Our evaluation results are shown in \cref{table:tracking_results} and in \cref{fig:auc}. Based on our experimentation, the top performing methods are the ones that employ a descriptor constancy assumption, namely: bit-planes and DF\@. However, bit-planes is more efficient and it performed significantly better for the `out of plane rotation', which induces perspective change in the image. In fact, all tested algorithms, with the exception of bit-planes, performed poorly with this data. Algorithms that use a robust function (ECC) and the ones that attempt to estimate illumination (DIC) performed well, but fell behind in comparison to descriptor constancy and gradient constraint.

\begin{table}
  \centering\small
  \scalebox{0.9}{%
  \begin{tabular}{rcccc}
 \toprule
    & \multicolumn{4}{c}{Template area} \\
    & $75 \times 57$ & $150 \times 115$ & $300 \times 230$ & $640\times 460$ \\ \midrule
    Intensity & $650$ & $360$ & $140$ & $45$ \\
    Bit-planes & $460$ & $170$ & $90$  & $35$ \\
    \bottomrule
  \end{tabular}}
  \caption{Plane template tracking runtime in frames per second (FPS) on a single core Intel i7-2460M CPU @ 2.8 Ghz\@.}\label{table:runtime}
\end{table}

\begin{table*}
  \centering\small
  \begin{tabular}{l cccccc}
\toprule%
    & \texttt{br} & \texttt{bu} & \texttt{mi} & \texttt{pa} & \texttt{su} & \texttt{wd} \\ \midrule
    & \multicolumn{6}{c}{Out of Plane Rotation}                                         \\ \cmidrule(r){2-7}
  BP  & \textbf{100.0} (\textcolor{green}{99.38}) & \textbf{100.0} (99.51) & \textbf{87.50} (99.38) & \textbf{97.92} (99.26) & \textbf{79.17} (99.57) & \textbf{93.75} (99.30) \\
 ECC  & 25.00 (96.16) & 33.33 (95.85) & 25.00 (95.99) & 33.33 (96.65) & 20.83 (95.52) & 18.75 (95.14)   \\
DIC-1  & 25.00 (96.20) & 33.33 (95.83) & 25.00 (95.98) & 33.33 (96.73) & 20.83 (95.95) & 18.75 (95.46)   \\
DIC-2  & 25.00 (96.22) & 35.42 (95.56) & 25.00 (95.51) & 35.42 (96.42) & 25.00 (96.22) & 18.75 (95.06)   \\
  DF  & 91.67 (99.51) & 93.75 (99.44) & 79.17 (99.70) & 85.42 (99.75) & 70.83 (99.60) & 83.33 (99.51)   \\
  GC  & \textbf{100.0} (99.24) & 95.83 (99.66) & 87.50 (99.52) & 93.75 (99.51) & 62.50 (98.88) & 91.67 (99.34)  \\
  GM  & 62.50 (99.86) & 83.33 (99.62) & 77.08 (99.72) & 77.08 (99.81) & 58.33 (99.71) & 62.50 (99.66)   \\
  LK  & 93.75 (99.68) & 91.67 (99.70) & 83.33 (99.32) & 91.67 (99.63) & 37.50 (97.64) & 66.67 (99.63)   \\ \midrule
     & \multicolumn{6}{c}{Dynamic Lighting Change} \\ \cmidrule(r){2-7}
BP   & \textbf{100.0} (98.97) & \textbf{100.0} (99.08) & \textbf{100.0} (\textcolor{green}{99.13}) & \textbf{100.0} (\textcolor{green}{98.91}) & \textbf{100.0} (\textcolor{green}{98.98}) & \textbf{100.0} (99.02) \\
 ECC  & 16.33 (98.03) & 19.39 (99.00) & \textbf{100.0} (98.64) & \textbf{100.0} (98.69) & \textbf{100.0} (97.30) & 67.35 (98.55)  \\
DIC-1  & \textbf{100.0} (98.40) & \textbf{100.0} (99.04) & \textbf{100.0} (98.77) & \textbf{100.0} (98.60) & 86.87 (96.02) & 20.41 (95.36) \\
DIC-2  & \textbf{100.0} (98.39) & \textbf{100.0} (98.85) & \textbf{100.0} (98.61) & \textbf{100.0} (98.58) & 85.86 (96.42) & 26.53 (97.73) \\
  DF  & \textbf{100.0} (\textcolor{green}{99.30}) & \textbf{100.0} (99.08) & \textbf{100.0} (98.35) & \textbf{100.0} (98.87) & 20.41 (99.36) & 68.37 (99.02) \\
  GC  & 17.35 (99.87) & \textbf{100.0} (\textcolor{green}{99.50}) & 22.45 (99.84) & 18.37 (99.88) & 12.24 (99.72) & 17.35 (99.84) \\
  GM  & 17.35 (98.99) & 19.39 (99.23) & 23.47 (99.10) & 19.39 (99.08) & 0.00 (0.00) & 0.00 (0.00) \\
  LK  & 13.27 (99.34) & 31.63 (98.26) & 18.37 (98.82) & 18.37 (99.32) & 12.24 (99.16) & 16.33 (98.96) \\ \midrule
      & \multicolumn{6}{c}{Static lighting change} \\ \cmidrule(r){2-7}
BP    & \textbf{100.0} (\textcolor{green}{99.76}) & \textbf{100.0} (\textcolor{green}{99.85}) & \textbf{100.0} (\textcolor{green}{99.61}) & \textbf{100.0} (\textcolor{green}{99.85}) & \textbf{100.0} (\textcolor{green}{99.63}) & \textbf{100.0} (99.76) \\
 ECC  & \textbf{100.0} (97.33) & \textbf{100.0} (97.67) & \textbf{100.0} (97.75) & \textbf{100.0} (97.41) & \textbf{100.0} (96.79) & \textbf{100.0} (97.55) \\
DIC-1  & \textbf{100.0} (97.70) & \textbf{100.0} (97.77) & \textbf{100.0} (97.80) & \textbf{100.0} (97.20) & 98.72 (96.58) & 89.74 (96.19) \\
DIC-2  & \textbf{100.0} (97.58) & 79.49 (97.59) & \textbf{100.0} (97.07) & \textbf{100.0} (97.13) & 89.74 (95.75) & 79.49 (96.38) \\
  DF  & \textbf{100.0} (99.68) & \textbf{100.0} (99.51) & 76.92 (99.71) & \textbf{100.0} (99.77) & 74.36 (99.70) & \textbf{100.0} (\textcolor{green}{99.83}) \\
  GC  & 74.36 (99.73) & 74.36 (99.84) & 48.72 (99.97) & 74.36 (99.76) & 48.72 (99.74) & 51.28 (99.88) \\
  GM  & 48.72 (99.88) & 74.36 (99.75) & 74.36 (99.66) & 74.36 (99.81) & 48.72 (99.76) & 48.72 (99.83) \\
  LK  & 48.72 (99.80) & 74.36 (99.67) & 48.72 (99.95) & 48.72 (99.93) & 48.72 (99.40) & 48.72 (99.94) \\ \bottomrule
\end{tabular}

    \caption{Template tracking evaluation~\cite{Gauglitz_IJCV2011}. We show the percentage of successfully tracked frames. In parenthesis we show the average percentage of overlap for all successfully tracked frames. The available textures are: \texttt{br} (bricks), \texttt{bu} (building), \texttt{mi} (mission), \texttt{pa} (paris), \texttt{su} (sunset), and \texttt{wd} (wood).}\label{table:tracking_results}
\end{table*}

\begin{figure*}
  \centering\small
  \begin{subfigure}{\linewidth}
  \centering
  \includegraphics[width=0.19\linewidth]{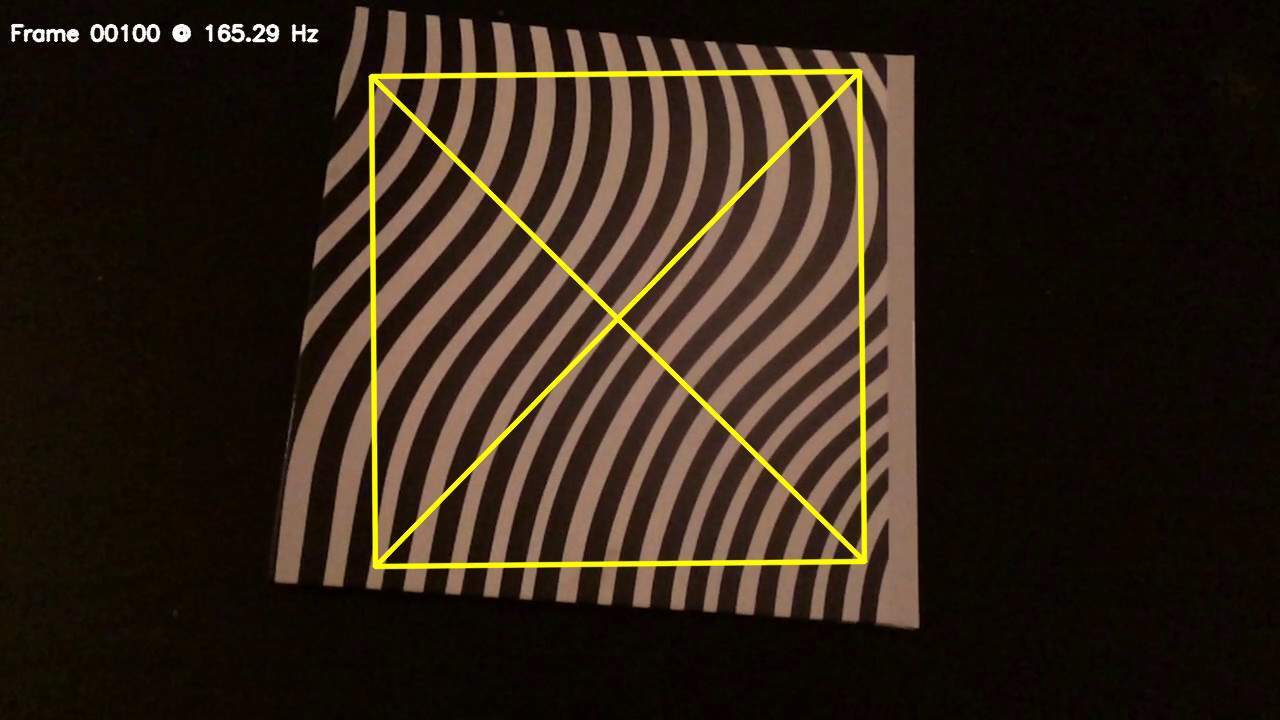}
  \includegraphics[width=0.19\linewidth]{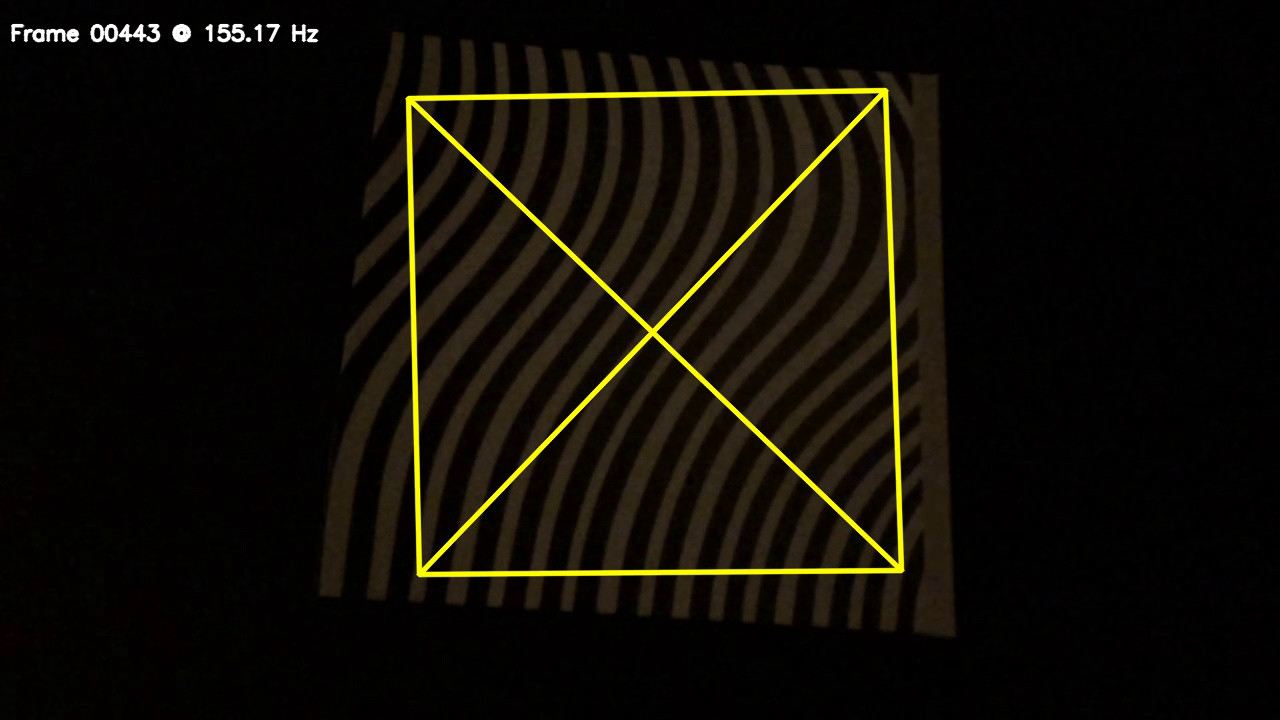}
  \includegraphics[width=0.19\linewidth]{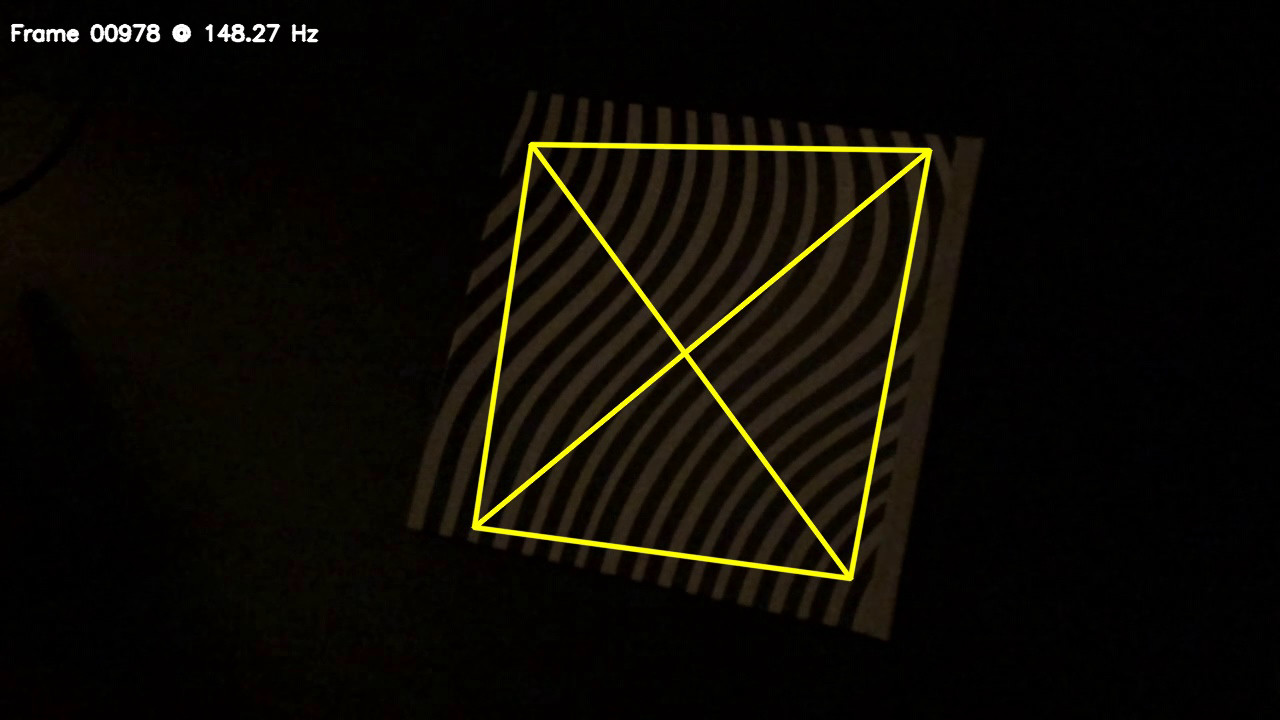}
  \includegraphics[width=0.19\linewidth]{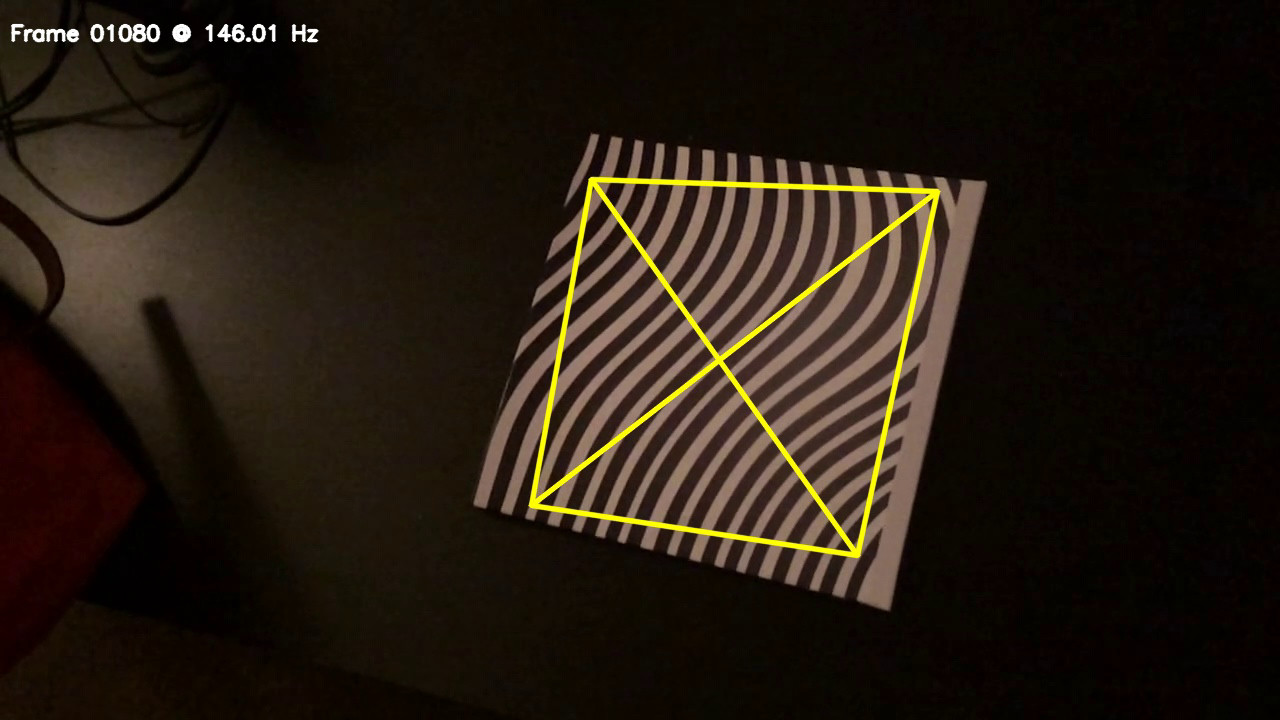}
  \includegraphics[width=0.19\linewidth]{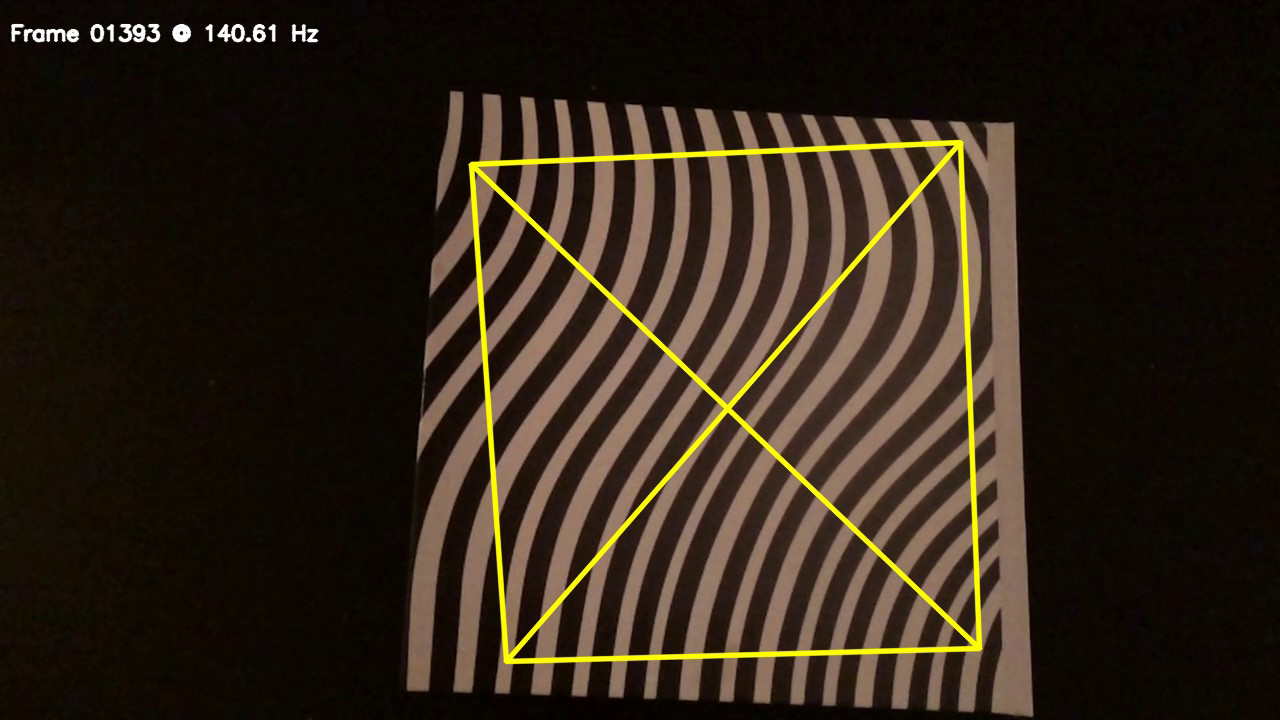}
  \caption{Sudden lighting change and ambiguous texture.}\label{fig:v8}
  \end{subfigure}
  \begin{subfigure}{\linewidth}
  \centering
  \includegraphics[width=0.19\linewidth]{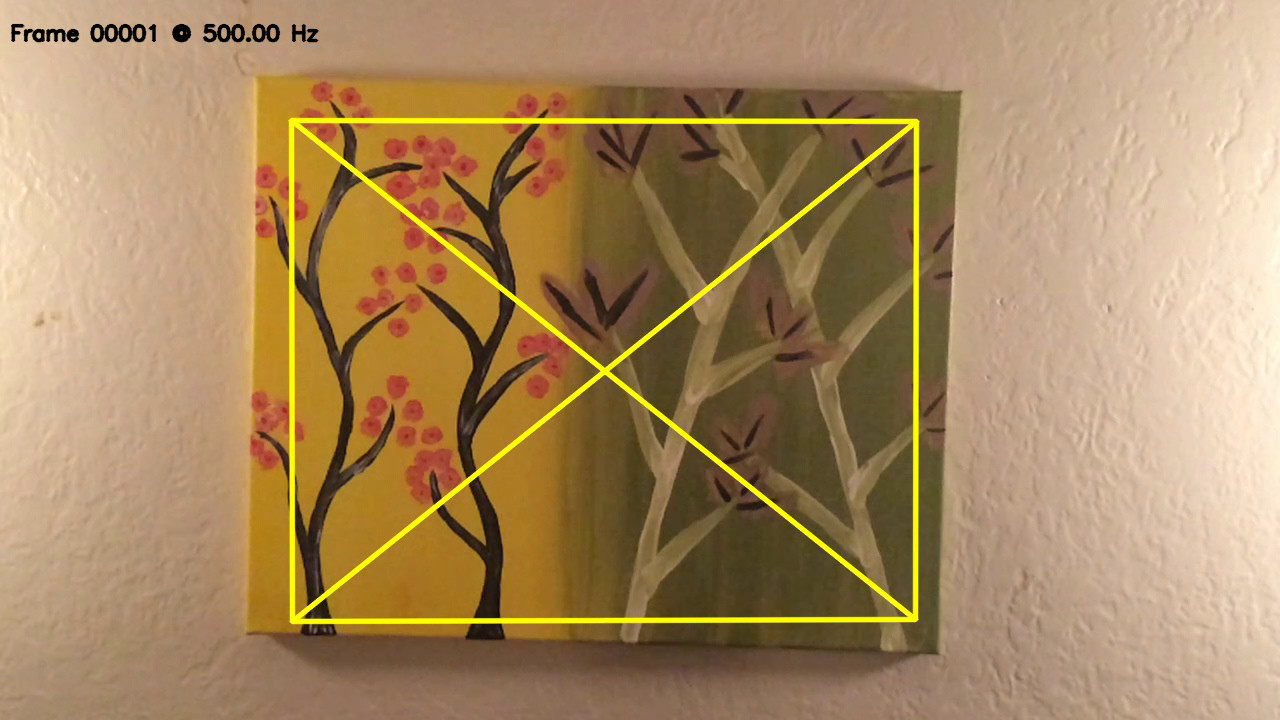}
  \includegraphics[width=0.19\linewidth]{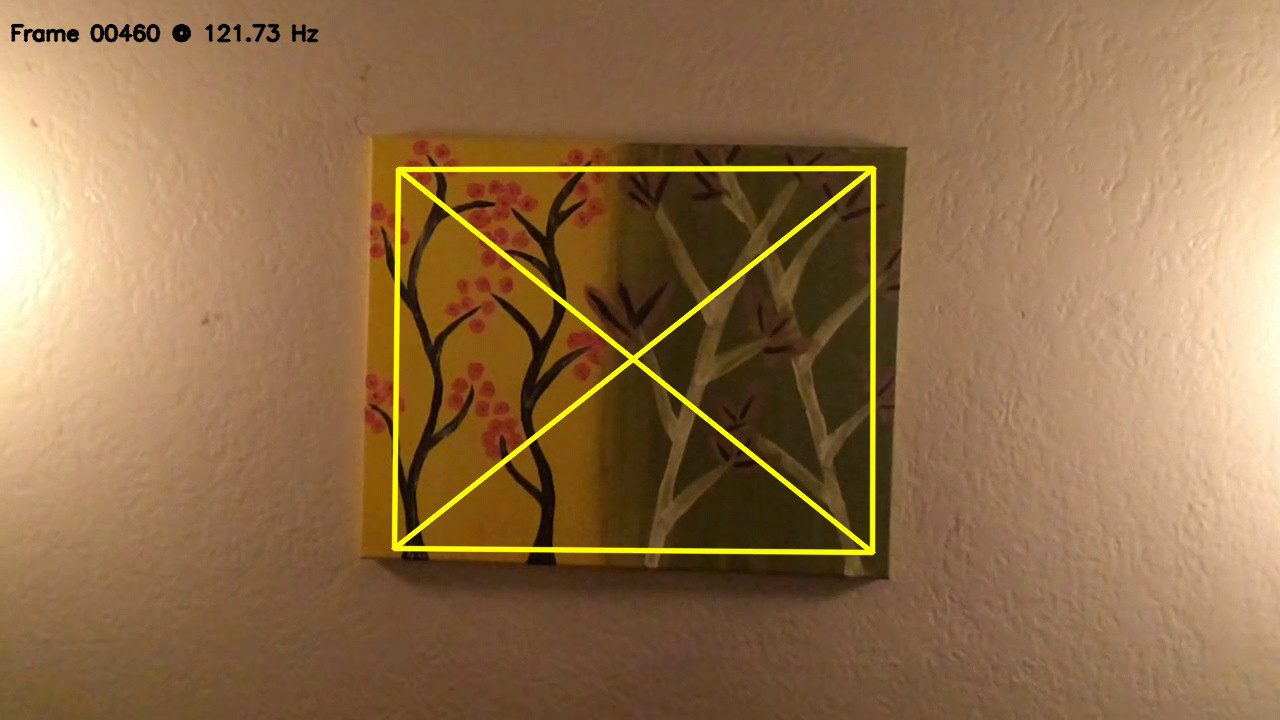}
  \includegraphics[width=0.19\linewidth]{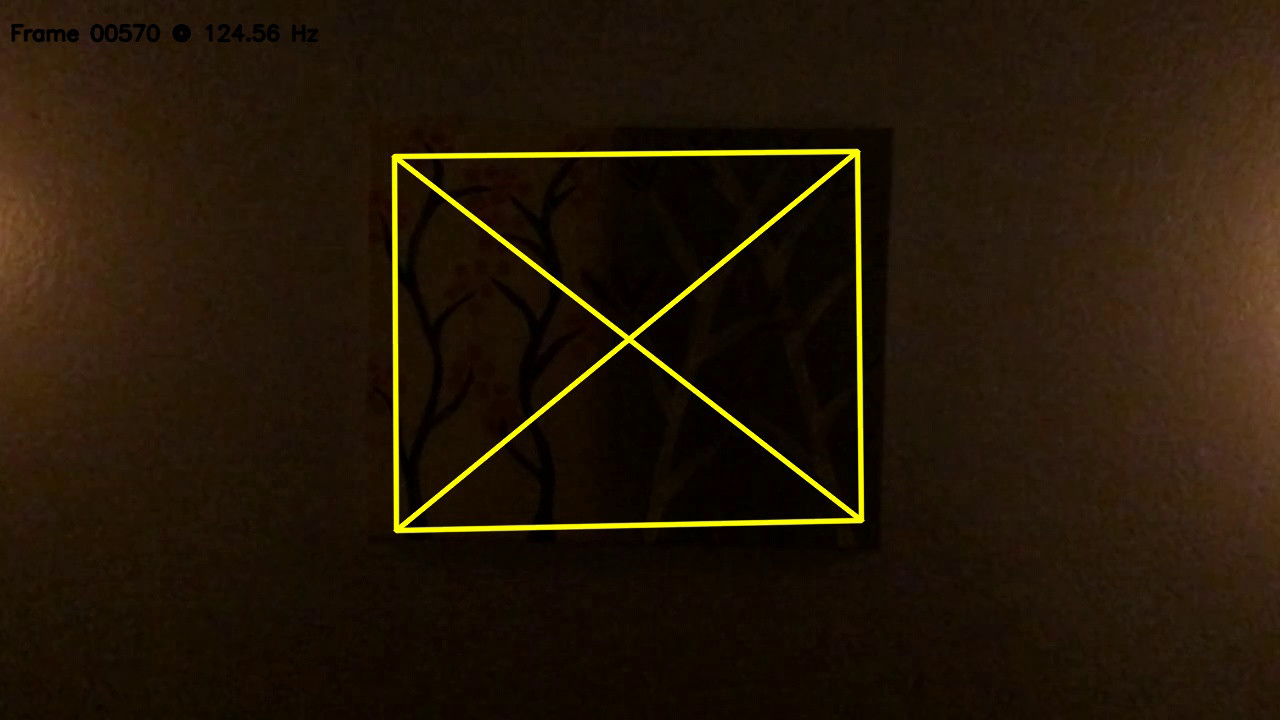}
  \includegraphics[width=0.19\linewidth]{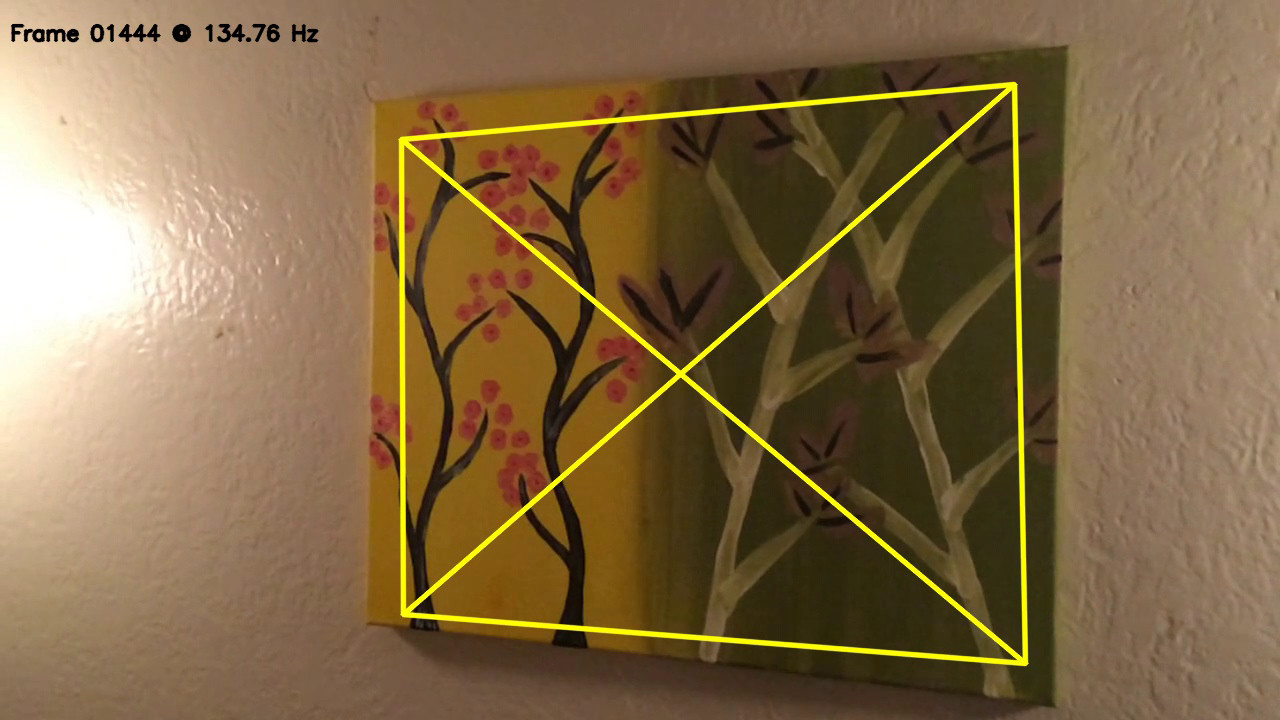}
  \includegraphics[width=0.19\linewidth]{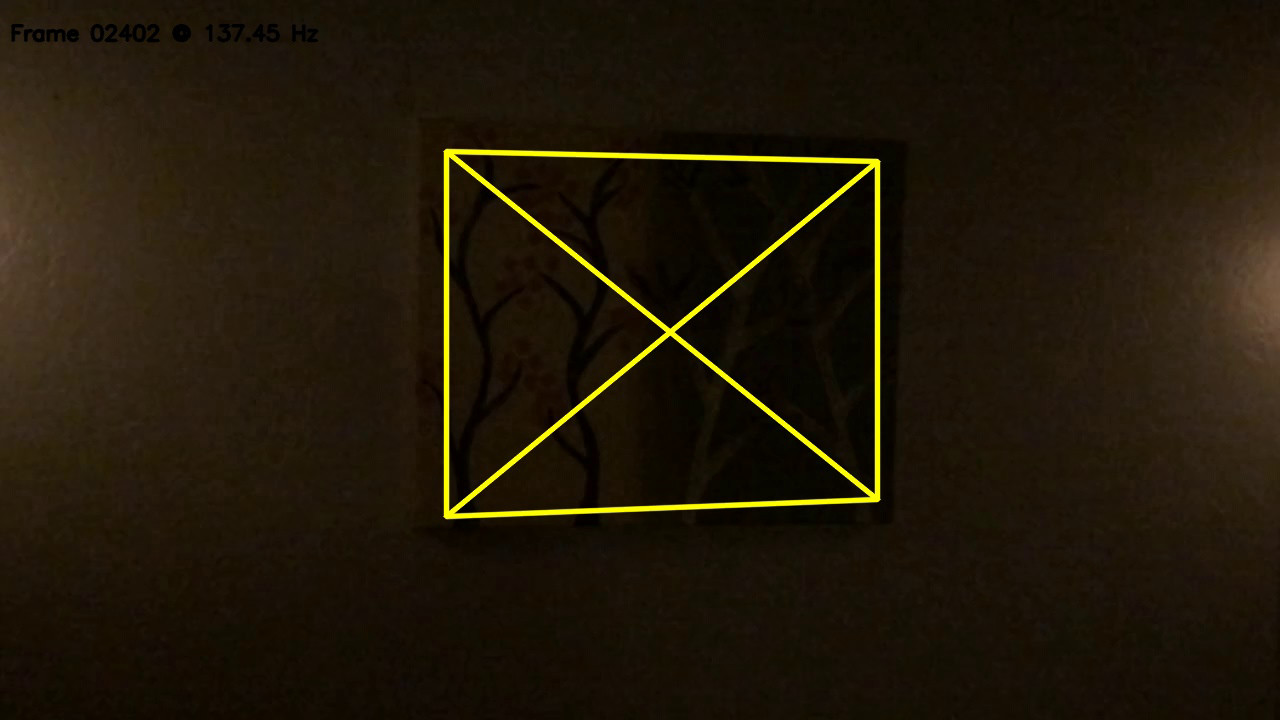}
  \caption{Sudden lighting change and perspective distortion with medium texture.}\label{fig:v7}
  \end{subfigure}
  \begin{subfigure}{\linewidth}
  \centering
  \includegraphics[width=0.19\linewidth]{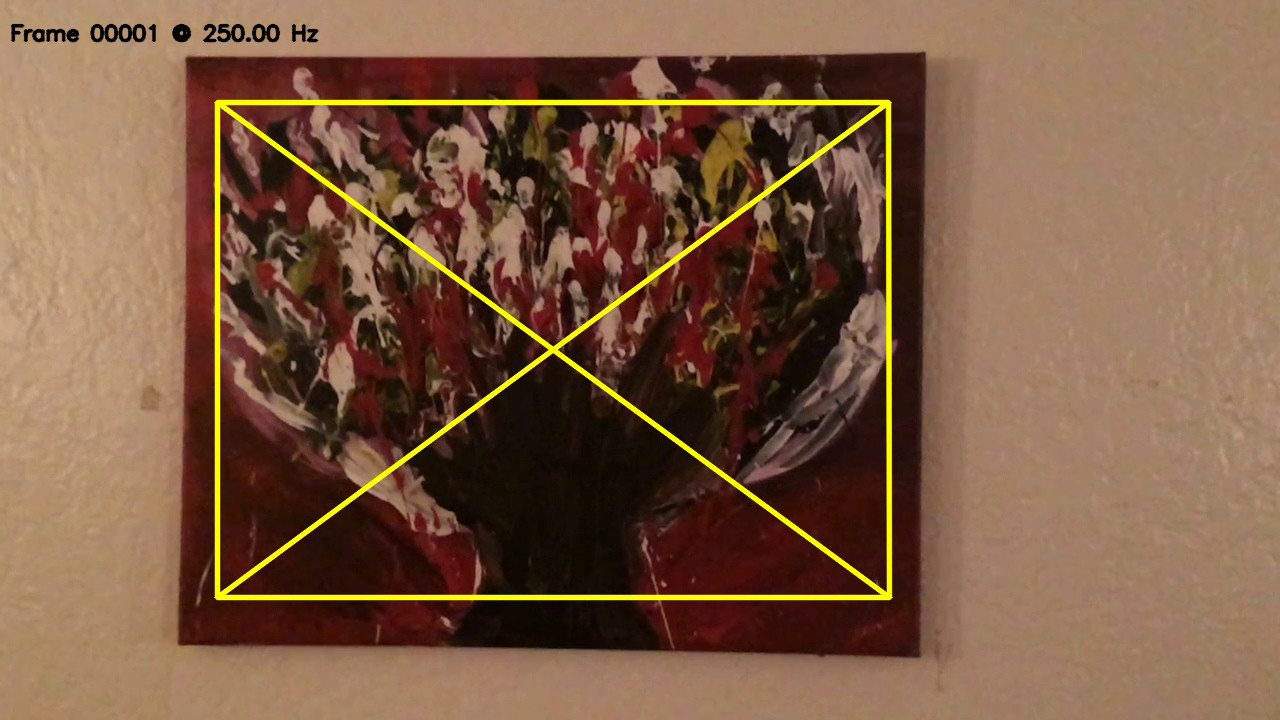}
  \includegraphics[width=0.19\linewidth]{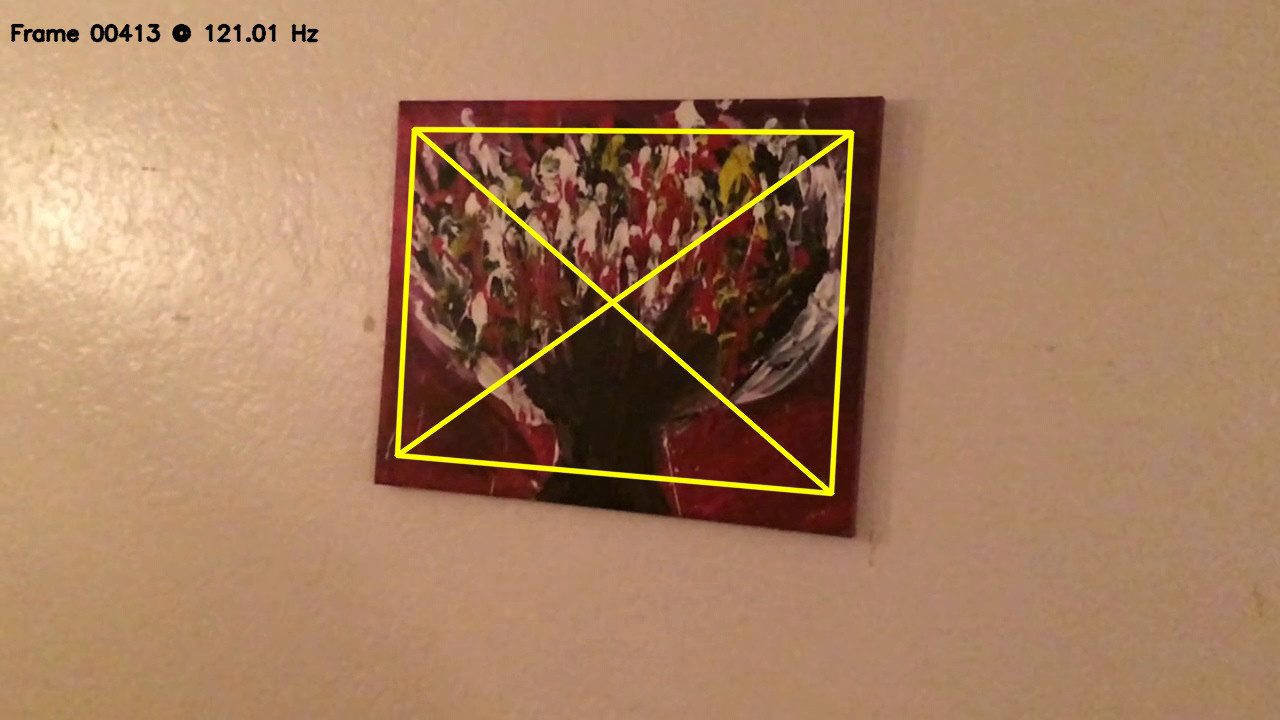}
  \includegraphics[width=0.19\linewidth]{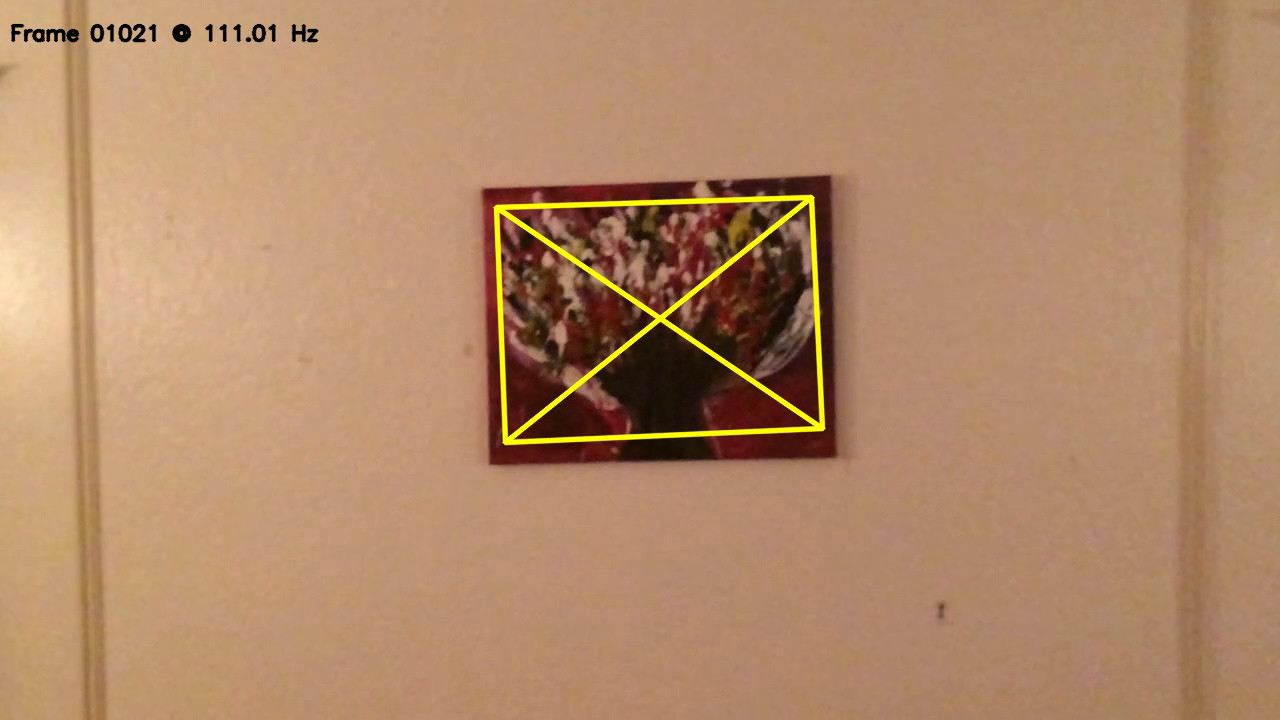}
  \includegraphics[width=0.19\linewidth]{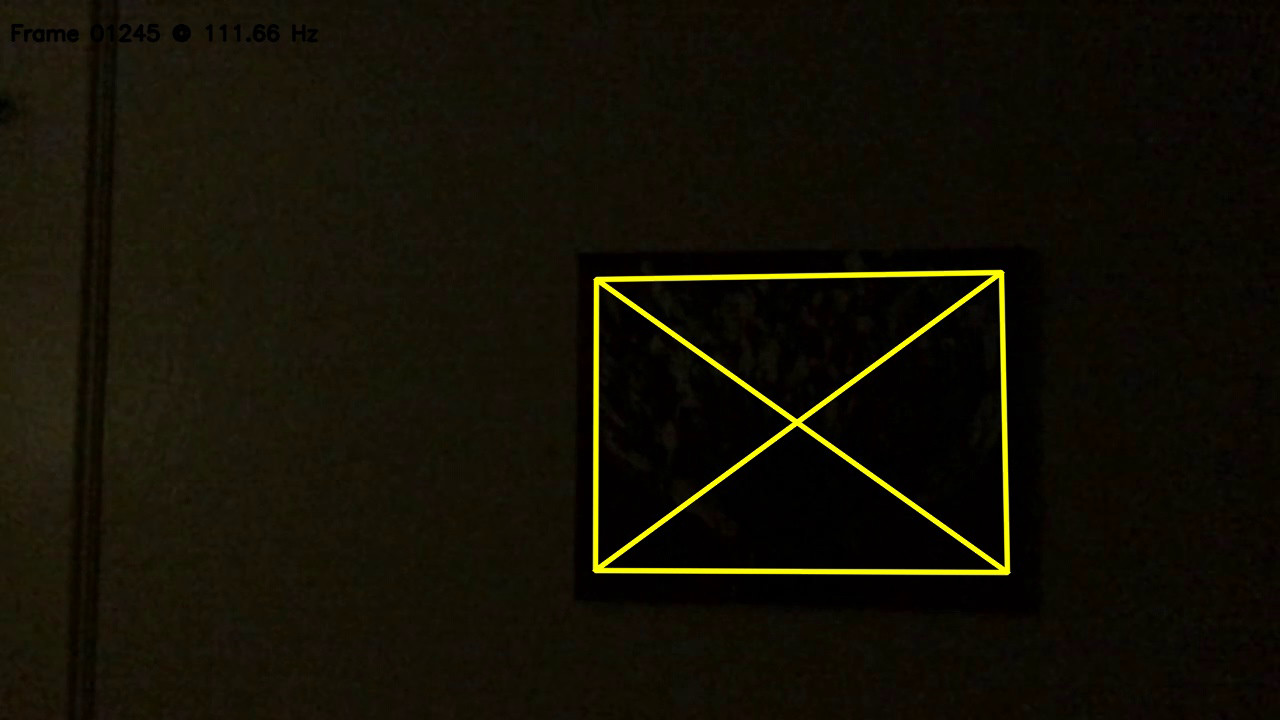}
  \includegraphics[width=0.19\linewidth]{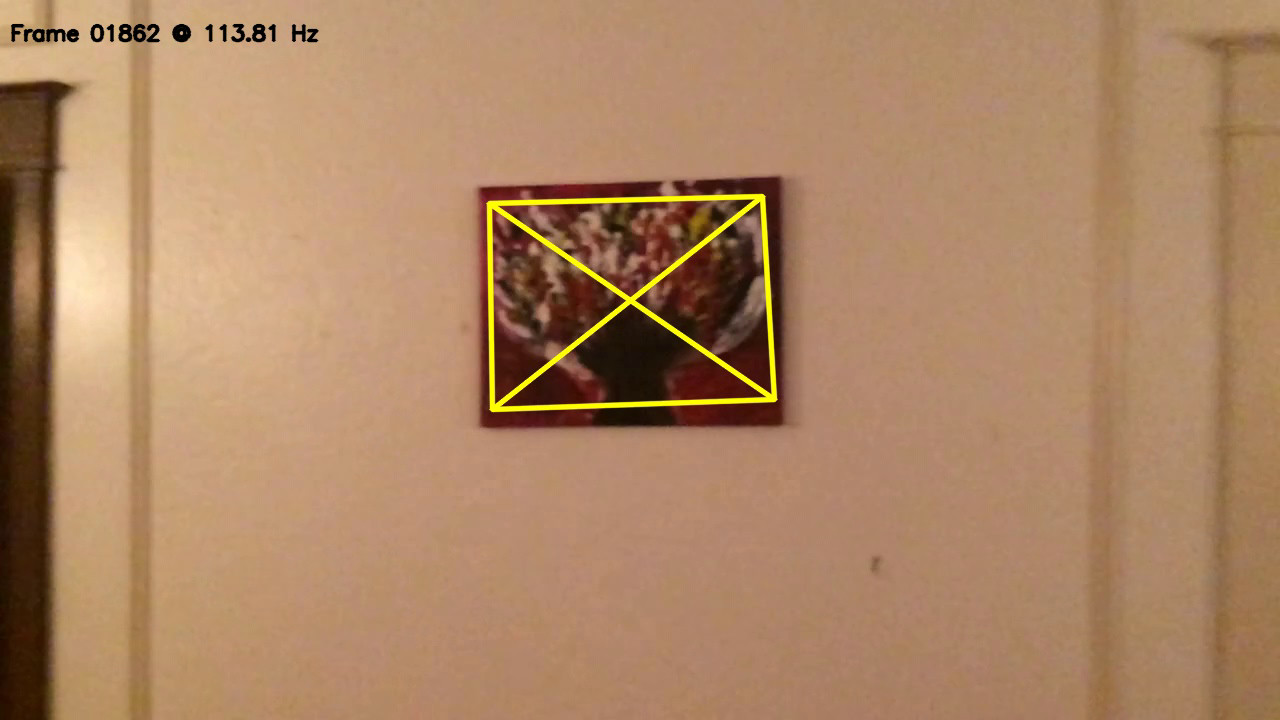}
  \caption{Sudden lighting change and motion blur with high texture.}\label{fig:v9}
  \end{subfigure}
  \caption{High frame rate data at $120$ Hz captured using an iPhone 5s. Dataset contains different textures under sudden lighting change, low lighting, and motion blur. All data and code will be publicly available for the research community.}\label{fig:slomo}
\end{figure*}

\begin{table}
\centering\small
\scalebox{0.9}{%
\begin{tabular}{r ccc ccc}
\toprule
   template & \multicolumn{3}{c}{iPad Air 2} & \multicolumn{3}{c}{iPhone 5s} \\
  size & BP & ORB & BRISK & BP & ORB & BRISK \\
  \cmidrule(l){1-1}
  \cmidrule(lr){2-4}
  \cmidrule(l ){5-7}
  $70\times 55$   & \textbf{123} & N/A & N/A &  \textbf{50} & N/A & N/A \\
  $150\times 115$ & \textbf{48}  & 15  & 15  &  \textbf{22} & 13  & 13 \\
  $311\times 230$ & \textbf{17}  & 12  & 14  &  \textbf{10} & 8   & 11 \\ \bottomrule
\end{tabular}}
\caption{Template tracking running time on ARM architecture using a single CPU core in frames per second (FPS). The bottleneck for bit-planes is image resizing and warping, which could be alleviated using the GPU\@. Results are averaged over three videos of challenging data totalling 6446 frames.}
\label{table:arm}
\end{table}

\subsection{Results on mobile devices}
We further evaluate the work on high frame rate data (Slo-mo) using two smart mobile devices: the iPad Air 2 and the iPhone 5s. In addition to compression artifacts, we made the data more challenging by turning off the lights multiples times to cause sudden lighting change and low illumination. The videos are recorded with unsteady hands causing further motion blur. An example of the videos is shown in \cref{fig:slomo} featuring an ambiguously textured object in \cref{fig:v8}, normal levels of texture in \cref{fig:v7} as well as higher amount of texture in \cref{fig:v9}. The first image in \cref{fig:slomo} shows the selected template, which we hold fixed throughout tracking. The total number of frames from the videos combined is $6447$.

We compare the performance of dense tracking using bit-planes with the RANSAC-based tracking by detection using two types of binary descriptors, ORB~\cite{orb} and BRISK~\cite{brisk}. In terms of efficiency, even though our mobile device implementation does not make use of NEON instructions or the GPU, we outperform opencv3's optimized implementations of ORB and BRISK by a substantial margin.

More importantly, our approach is more robust. Feature-based tracking failed approximately on $15\%$ of the frames due to either: (i) inability to detect features under low light, and (ii) RANSAC failure due to imprecise correspondences under motion blur.

Perhaps more interestingly, bit-planes is able to perform well and improve efficiency by reducing image resolution. In fact, tracking speed more than doubles when reducing the template size by half. However, this is not the case with sparse features as memory overhead depends on the number of extracted keypoints, which we kept fixed at $512$. It is possible to improve the tracking speed of ORB and BRISK by reducing the number of extracted keypoints. However, lowering the number of keypoints must be done carefully as not to compromise the robustness of the system. We note that the ability to work with lower resolution is important on mobile devices, not only for efficiency considerations, but also for power consumption.

Finally, we note that while dense bit-planes tracking produces faster and more accurate results, its main limitation is the inability to recover if the template is lost due to occlusions or significant drift. In such cases, track by detection can be of immense value to re-initialize LK-based methods if needed.

\section{Conclusions}
In this work, we proposed a multi-channel representation that enables nonlinear gradient-based optimization algorithms to work with binary features. We arrive at the surprising result that binary data is suitable for gradient-based optimization, as the local approximation of gradients per channel is well-approximated with a quadratic form.

We used the multi-channel representation in a Lucas \& Kanade (LK) image alignment framework with our proposed bit-planes descriptors, which greatly improves the robustness to arbitrary illumination variations without significantly increasing computational demands. In addition, we obtained a precise subpixel localization of binary descriptors at speeds faster than real time.

In the context of binary features, least squares minimization over the multi-channel representation is equivalent to minimizing the Hamming distance. Hence, we are able to minimize the Hamming distance in a standard least squares optimization framework.
{%
  \small
  \bibliographystyle{ieee}
  \bibliography{bib}
}
\end{document}